\documentclass{article}
\usepackage{ijcai19}

\usepackage{times}
\usepackage{soul}
\usepackage{url}
\usepackage[utf8]{inputenc}
\usepackage[small]{caption}
\usepackage{graphicx}
\usepackage{amsmath}
\usepackage{booktabs}
\usepackage{algorithm}
\usepackage{algorithmic}
\usepackage{enumerate}
\usepackage{multirow}
\usepackage{array}
\title{Vision-based Robotic Grasping From Object Localization, Object Pose Estimation to Grasp Estimation for Parallel Grippers: A Review}

\author{Guoguang Du \and Kai Wang \and Shiguo Lian \and Kaiyong Zhao
\affiliations
CloudMinds Technologies
\emails
george.du@cloudminds.com
}

\begin{document}

\maketitle

\begin{abstract}
  This paper presents a comprehensive survey on vision-based robotic grasping. We conclude three key tasks during vision-based robotic grasping, which are object localization, object pose estimation and grasp estimation. In detail, the object localization task contains object localization without classification, object detection and object instance segmentation. This task provides the regions of the target object in the input data. The object pose estimation task mainly refers to estimating the 6D object pose and includes correspondence-based methods, template-based methods and voting-based methods, which affords the generation of grasp poses for known objects. The grasp estimation task includes 2D planar grasp methods and 6DoF grasp methods, where the former is constrained to grasp from one direction. These three tasks could accomplish the robotic grasping with different combinations. Lots of object pose estimation methods need not object localization, and they conduct object localization and object pose estimation jointly. Lots of grasp estimation methods need not object localization and object pose estimation, and they conduct grasp estimation in an end-to-end manner. Both traditional methods and latest deep learning-based methods based on the RGB-D image inputs are reviewed elaborately in this survey. Related datasets and comparisons between state-of-the-art methods are summarized as well. In addition, challenges about vision-based robotic grasping and future directions in addressing these challenges are also pointed out.
\end{abstract}

\section{Introduction}
\label{sec:1}

An intelligent robot is expected to perceive the environment and interact with it. Among the essential abilities, the ability to grasp is fundamental and significant in that it will bring enormous power to the society~\cite{2018RoboticApplications}. For example, industrial robots can accomplish the pick-and-place task which is laborious for human labors, and domestic robots are able to provide assistance to disabled or elder people in their daily grasping tasks. Endowing robots with the ability to perceive has been a long-standing goal in computer vision and robotics discipline.

\begin{figure}[!t]
\centering
\includegraphics[scale=0.31]{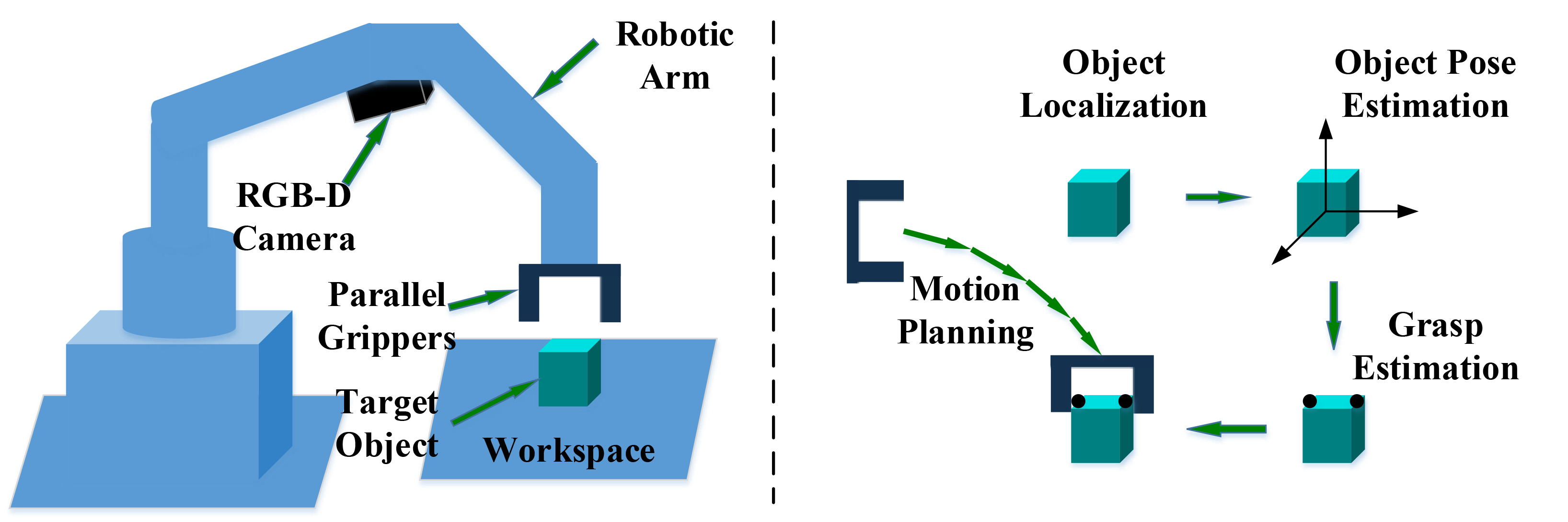}
\caption{The grasp detection system. (Left)The robotic arm, equipped with one RGB-D camera and one parallel gripper, is to grasp the target object placed on a planar work surface. (Right)The grasp detection system involves target object localization, object pose estimation, and grasp estimation.}
\label{fig:GraspDetectionSystem}
\end{figure}

As much as being highly significant, robotic grasping has long been researched. The robotic grasping system~\cite{2017RoboticGrasp} is considered as being composed of the following sub-systems: the grasp detection system, the grasp planning system and the control system. Among them, the grasp detection system is the key entry point, as illustrated in Fig.~\ref{fig:GraspDetectionSystem}. The grasp planning system and the control system are more relevant to the motion and automation discipline, and in this survey, we only concentrate on the grasp detection system.

\begin{figure}[!t]
\centering
\includegraphics[scale=0.5]{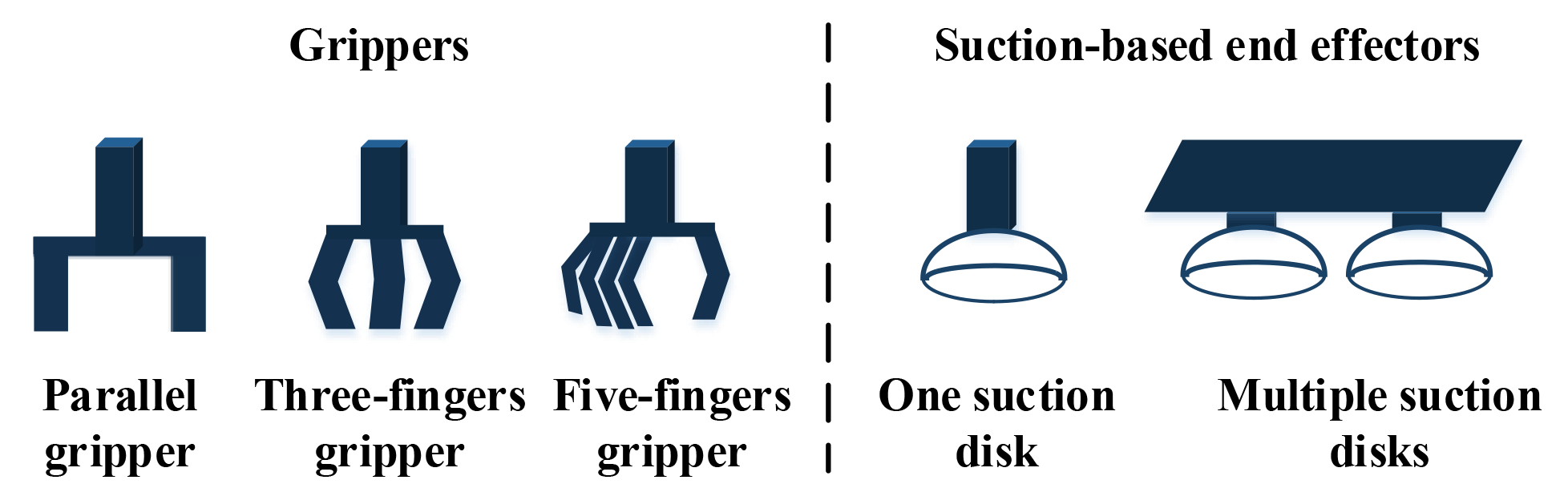}
\caption{Different kinds of end effectors. (Left)Grippers. (Right)Suction-based end effectors. In this paper, we mainly consider parallel grippers.}
\label{fig:EndEffectors}
\end{figure}

The robotic arm and the end effectors are essential components of the grasp detection system. Various 5-7 DoF robotic arms are produced to ensure enough flexibilities and they are equipped on the base or a human-like robot. Different kinds of end effectors, such as grippers and suction disks, can achieve the object picking task, as shown in Fig.~\ref{fig:EndEffectors}. The majority of methods paid attentions on parallel grippers~\cite{2017DexNet2,2017MultiViewSelf}, which is a relatively simple situation. With the struggle of academia, dexterous grippers~\cite{2019GeneratingHighDoF,2019FfficientGraspMultiFinger,2019SolvingRubicCube} are researched to accomplish complex grasp tasks. In this paper, we only talk about grippers, since suction-based end effectors are relatively simple and limited in grasping complex objects. In addition, we concentrate on methods using parallel grippers, since this is the most widely researched.

The essential information to grasp the target object is the 6D gripper pose in the camera coordinate, which contains the 3D gripper position and the 3D gripper orientation to execute the grasp. The estimation of 6D gripper poses varies aiming at different grasp manners, which can be divided into the 2D planar grasp and the 6DoF grasp.

2D planar grasp means that the target object lies on a planar workspace and the grasp is constrained from one direction. In this case, the height of the gripper is fixed and the gripper direction is perpendicular to one plane. Therefore, the essential information is simplified from 6D to 3D, which are the 2D in-plane position and 1D rotation angle. In earlier years when the depth information is not easily captured, the 2D planar grasp is mostly researched. The mostly used scenario is to grasp machine components in the factory. The grasping contact points are evaluated whether they can afford the force closure~\cite{1993FindingAntipodal}. With the development of deep learning, large number of methods treated oriented rectangles as the grasp configuration, which could be beneficial from the mature 2D detection frameworks. Since then, the capabilities of 2D planar grasp are enlarged extremely and the target objects to be grasped are extended from known objects to novel objects. Large amounts of methods by evaluating the oriented rectangles~\cite{2011EfficientGrasping,2015DeepLearnDetect,2016SupersizingSelfSuper,2017DexNet2,2018ClassificationBasedGrasp,2015RealGrasp,2017RobustRobotGrasp,2017RoboticGrasp,2018RealWorldMultiObj,2018RealTimeHighly,2018FullyConvolutionalGrasp} are proposed. Besides, some deep learning-based methods of evaluating grasp contact points~\cite{2018RoboticPick,2019MetaGrasp,2018ClosingTheLoop} are also proposed in recent years.

\begin{figure}[!t]
\centering
\includegraphics[scale=0.42]{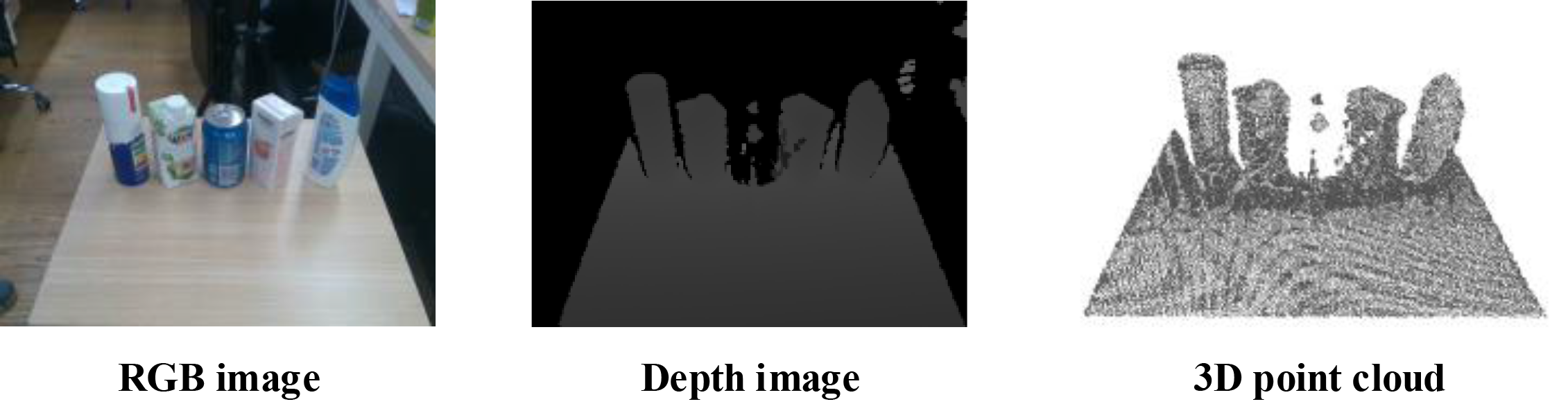}
\caption{A RGB-D image. The depth image is transformed into 3D point cloud.}
\label{fig:depthimage}
\end{figure}

6DoF grasp means that the gripper can grasp the object from various angles in the 3D space, and the essential 6D gripper pose could not be simplified. In early years, analytical methods were utilized to analyze the geometric structure of the 3D data, and the points suitable to grasp were found according to force closure. Sahbani et al.~\cite{2012GraspSurvey} presented an overview of 3D object grasping algorithms, where most of them deal with complete shapes. With the development of sensor devices, such as Microsoft Kinect, Intel RealSense, etc, researchers can obtain the depth information of the target objects easily and modern grasp systems are equipped with RGB-D sensors, as shown in Fig.~\ref{fig:depthimage}. The depth image can be easily lifted into 3D point cloud with the camera intrinsic parameters and the depth image-based 6DoF grasp becomes the hot research areas. Among 6DoF grasp methods, most of them aim at known objects where the grasps could be precomputed, and the problem is thus transformed into a 6D object pose estimation problem~\cite{2019DenseFusion,2020LRF-Net,20206DoFViaDPVL,2020PVN3D}. With the development of deep learning, lots of methods~\cite{2017GPD,2019Pointnetgpd,20196DoFGraspNet,2019S4g,2020REGNet} illustrated powerful capabilities in dealing with novel objects.

Both 2D planar grasp and 6DoF grasp contain common tasks which are object localization, object pose estimation and grasp estimation.

In order to compute the 6D gripper pose, the first thing to do is to locate the target object. Aiming at object localization, there exist three different situations, which are object localization without classification, object detection and object instance segmentation. Object localization without classification means obtaining the regions of the target object without classifying its category. There exist cases that the target object could be grasped without knowing its category. Object detection means detecting the regions of the target object and classifying its category. This affords the grasping of specific objects among multiple candidate objects. Object instance segmentation refers to detecting the pixel-level or point-level instance objects of a certain class. This provides delicate information for pose estimation and grasp estimation. Early methods assume that the object to grasp is placed in a clean environment with simple background and thus simplifies the object localization task, while in relatively complex environments their capabilities are quite limited. Traditional object detection methods utilized machine learning methods to train classifiers based on hand-crafted 2D descriptors. However, these classifiers show limited performance since the limitations of hand-crafted descriptors. With the deep learning, the 2D detection and 2D instance segmentation capabilities improves a lot, which affords object detection in more complex environments.

\begin{figure*}[htbp]
\centering
\includegraphics[scale=0.45]{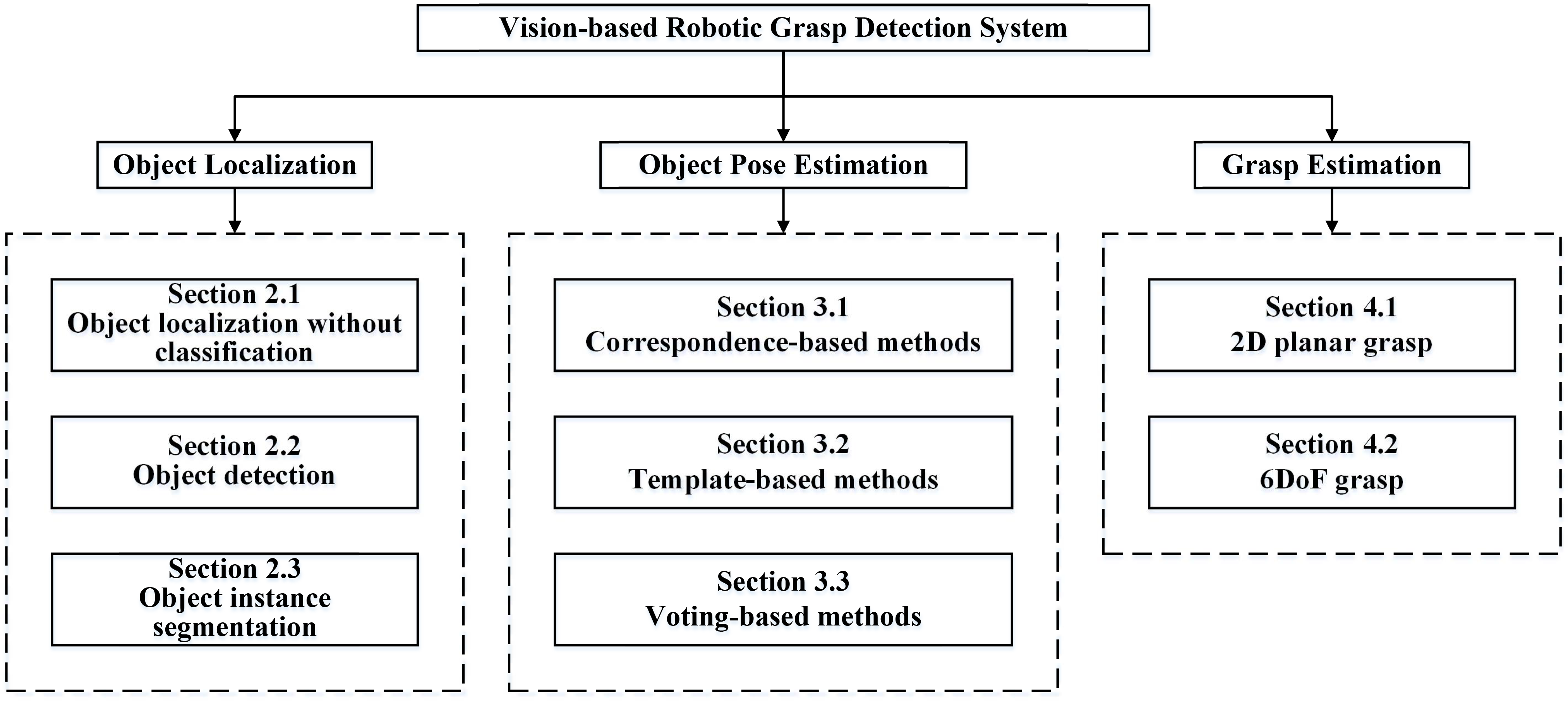}
\caption{A taxonomy of tasks in vision-based robotic grasp detection system.}
\label{fig:WholeTaxonomy}
\end{figure*}

Most of the current robotic grasping methods aim at known objects, and estimating the object pose is the most accurate and simplest way to a successful grasp. There exist various methods in computing the 6D object poses, which varies from 2D inputs to 3D inputs, from traditional methods to deep learning methods, from textured objects to textureless or occluded objects. In this paper, we categorize these methods into correspondence-based methods, template-based methods and voting-based methods, where only feature points, the whole input and each meta unit are involved in computing the 6D object pose. Early methods tackled this problem in 3D domain by conducting partial registration. With the development of deep learning, methods using RGB image only can provide relatively high accurate 6D object poses, which highly improves the grasp capabilities.

Grasp estimation is conducted when we have the localized target object. Aiming at 2D planar grasp, the methods are divided into methods of evaluating the grasp contact points and methods of evaluating the oriented rectangles. Aiming at 6DoF grasp, the methods are categorized into methods based on the partial point cloud  and methods based on the complete shape. Methods based on the partial point cloud mean that we do not have the identical 3D model of the target object. In this case, two kinds of methods exist which are methods of estimating grasp qualities of candidate grasps and methods of transferring grasps from existing ones. Methods based on complete shape means that the grasp estimation is conducted on a complete shape. When the target object is known, the 6D object pose could be computed. When the target shape is unknown, it can be reconstructed from single-view point clouds, and grasp estimation could be conducted on the reconstructed complete 3D shape. With the joint development of the above aspects, the kinds of objects that could be grasped, the robustness of the grasp and the affordable complexity of the grasp scenario all have improved a lot, which affords many more applications in industrial as well as domestic applications.

Aiming at these tasks mentioned above, there have been some works~\cite{2012GraspSurvey,2014DataSurvey,2018ReviewSurvey} concentrating on one or a few tasks, while there is still lack of a comprehensive introduction on these tasks. These tasks are reviewed elaborately in this paper, and a taxonomy of these tasks is shown in Fig.~\ref{fig:WholeTaxonomy}. To the best of our knowledge, this is the first review that broadly summarizes the progress and promises new directions in vision-based robotic grasping. We believe that this contribution will serve as an insightful reference to the robotic community.

The remainder of the paper is arranged as follows. Section 2 reviews the methods for object localization. Section 3 reviews the methods for 6D object pose estimation. Section 4 reviews the methods for grasp estimation. The related datasets, evaluation metrics and comparisons are also reviewed in each section. Finally, challenges and future directions are summarized in Section 5.

\section{Object localization}
\label{sec:2}

Most of the robotic grasping approaches require the target object's location in the input data first. This involves three different situations: object localization without classification, object detection and object instance segmentation. Object localization without classification only outputs the potential regions of the target objects without knowing their categories. Object detection provides bounding boxes of the target objects as well as their categories. Object instance segmentation further provides the pixel-level or point-level regions of the target objects along with their categories.

\subsection{Object localization without classification}
\label{sec:2-1}

In this situation, the task is to find potential locations of the target object without knowing the category of the target object. There exist two cases: if you known the concrete shapes of the target object, you can fit primitives to obtain the locations. If you can not ensure the shapes of the target object, salient object detection(SOD) could be conducted to find the salient regions of the target object. Based on 2D or 3D inputs, the methods are summarized in Table~\ref{tab:localization}.

\begin{table*}[htbp]
\centering
\caption{Methods of object localization without classification.}
    \begin{tabular}{ m{1.5cm} m{5.5cm} m{9.0cm} }
    \hline
    Methods & Fitting shape primitives & Salient object detection \\
    \hline
        2D localization & Fitting ellipse~\cite{1996FittingEllipse}, Fitting polygons~\cite{1973FittingPolygons} & Jiang et al.~\cite{2013SalientObjectDetection}, Zhu et al.~\cite{2014SaliencyOptimization}, Peng et al.~\cite{2016SalientODVia}, Cheng et al.~\cite{2014GlobalContrast}, Wei et al.~\cite{2012GeodesicSaliency}, Shi et al.~\cite{2015Hierarchical}, Yang et al.~\cite{2013SaliencyDetectionVia}, Wang et al.~\cite{2016CorrespondenceDriven}, Guo et al.~\cite{2017VideoSaliency}, Zhao et al.~\cite{2015SaliencyDetectionByMulti}, Zhang et al.~\cite{2016UnconstrainedSalient}, DHSNet~\cite{2016DHSNet}, Hou et al.~\cite{2017DeeplySupervisedSOD}, PICANet~\cite{2018PICANet}, Liu et al.~\cite{2019EmployingDeepPart}, Qi et al.~\cite{2019MultiScaleCapsule} \\
    \hline
        3D localization & Rabbani et al.~\cite{2005EfficientHough}, Rusu et al.~\cite{2009CloseRangeScene}, Goron et al.~\cite{2012RobustlySegmenting}, Jiang et al.~\cite{2013LinearApproach}, Khan et al.~\cite{2015SeparatingObjects}, Zapata-Impata et al.~\cite{2019FastGeometry} & Peng et al.~\cite{2014RgbdSOD}, Ren et al.~\cite{2015ExploitingGlobal}, Qu et al.~\cite{2017RgbdSalientOD}, Han et al.~\cite{2018AdvancedDLSOD}, Chen et al.~\cite{2019MultiModelFusionSOD,2019CnnRGBDSOD}, Chen and Li~\cite{2018ProgressivelyComplementarity}, Piao et al.~\cite{2019DepthInduced}, Kim et al.~\cite{2008SegmentationSalientRegions}, Bhatia et al.~\cite{2013SegmentingSOD3D}, Pang et al.~\cite{2020HierarchicalDynamicFN} \\
    \hline
    \end{tabular}
\label{tab:localization}
\end{table*}

\subsubsection{2D localization without classification}
\label{sec:2-1-1}

\begin{figure}[htbp]
\centering
\includegraphics[scale=0.36  ]{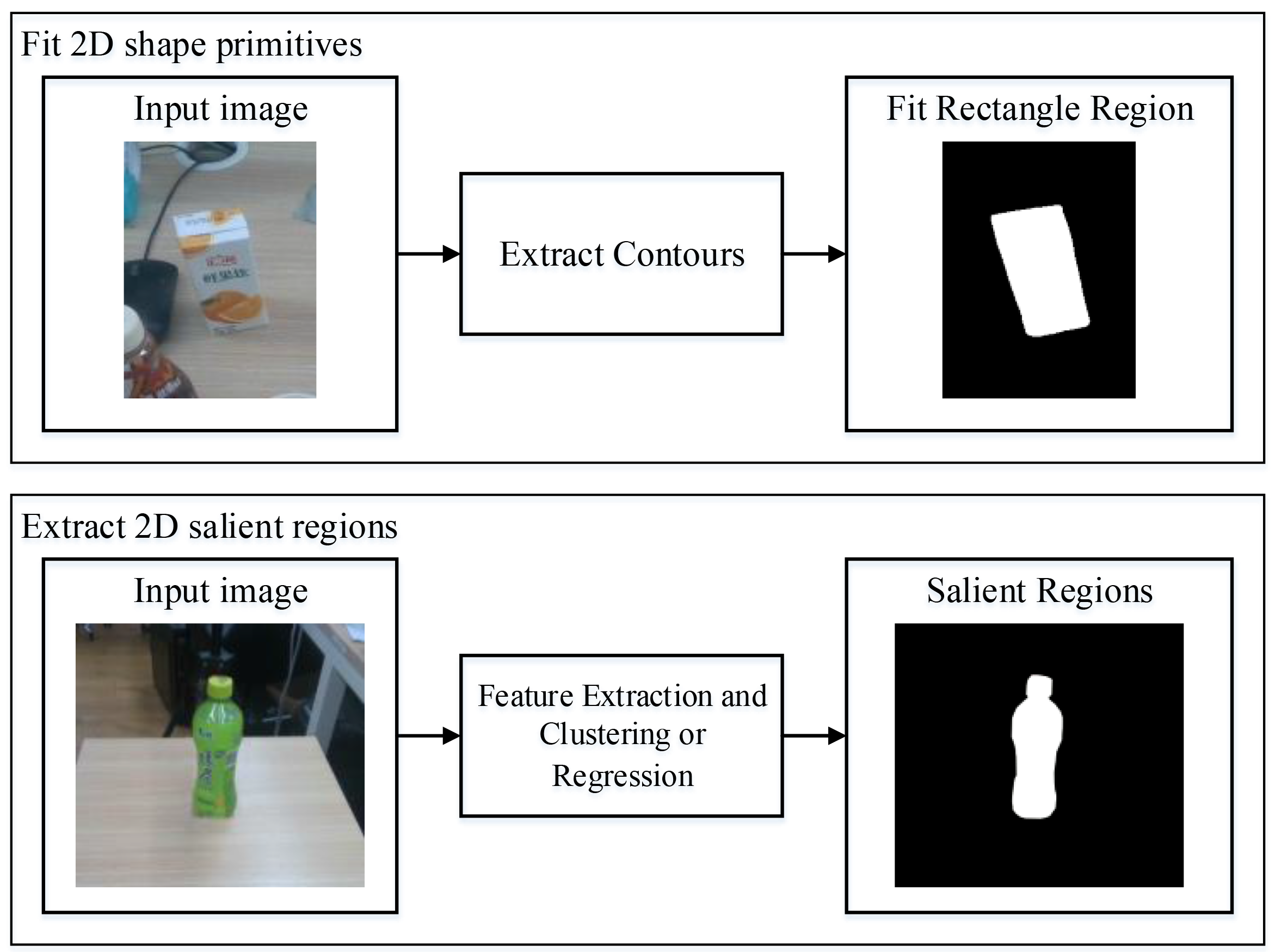}
\caption{Typical functional flow-chart of 2D object localization without classification.}
\label{fig:2dlocalization}
\end{figure}

This kind of methods deal with 2D image inputs, which are usually RGB images. According to whether the object's contour shape is known or not, methods can be divided into methods of fitting shape primitives and methods of salient object detection. Typical functional flow-chart of 2D object localization without classification is illustrated in Fig.~\ref{fig:2dlocalization}.

\paragraph{Fitting 2D shape primitives}

The shape of the target object could be an eclipse, a polygon or a rectangle, and these shapes could be regarded as shape primitives. Through fitting methods, the target object could be located. General procedures of this kind of methods usually contain enclosed contour extraction and primitive fitting. There exist many algorithms integrated in OpenCV~\cite{2008LearningOpenCV} for primitives fitting, such as fitting ellipse~\cite{1996FittingEllipse} and fitting polygons~\cite{1973FittingPolygons}. This kind of methods are usually used in 2D planar robotic grasping tasks, where the object are viewed from a fixed angle, and the target object are constrained with some known shapes.

\paragraph{2D salient object detection}

Compared with shape primitives, salient object regions could be represented in arbitrary shapes. 2D salient object detection(SOD) aims to locate and segment the most visually distinctive object regions in a given image, which is more like a segmentation task without object classification. Non-deep learning SOD methods exploit low-level feature representations~\cite{2013SalientObjectDetection,2014SaliencyOptimization,2016SalientODVia} or rely on certain heuristics such as color contrast~\cite{2014GlobalContrast}, background prior~\cite{2012GeodesicSaliency}. Some other methods conduct an over-segmentation process that generates regions~\cite{2015Hierarchical}, super-pixels~\cite{2013SaliencyDetectionVia,2016CorrespondenceDriven}, or object proposals~\cite{2017VideoSaliency} to assist the above methods.

Deep learning-based SOD methods have shown superior performance over traditional solutions since 2015. Generally, they can be divided into three main categories, which are Multi-Layer Perceptron (MLP)-based methods, Fully Convolutional Network (FCN)-based methods and Capsule-based methods. MLP-based methods typically extract deep features for each processing unit of an image to train an MLP-classifier for saliency score prediction. Zhao et al.~\cite{2015SaliencyDetectionByMulti} proposed a unified multi-context deep learning framework which involves global context and local context, which are fed into an MLP for foreground/background classification to model saliency of objects in images. Zhang et al.~\cite{2016UnconstrainedSalient} proposed a salient object detection system which outputs compact detection windows for unconstrained images, and a maximum a posteriori (MAP)-based subset optimization formulation for filtering bounding box proposals. The MLP-based SOD methods cannot capture well critical spatial information and are time-consuming. Inspired by Fully Convolutional Network (FCN)~\cite{2015FullyCN}, lots of methods directly output whole saliency maps. Liu and Han~\cite{2016DHSNet} proposed an end-to-end saliency detection model called DHSNet, which can simultaneously refine the coarse saliency map. Hou et al.~\cite{2017DeeplySupervisedSOD} introduced short connections to the skip-layer structures, which provides rich multi-scale feature maps at each layer. Liu et al.~\cite{2018PICANet} proposed a pixel-wise contextual attention network called PiCANet, which generates an attention map for each pixel and each attention weight corresponds to the contextual relevance at each context location. With the raise of Capsule Network~\cite{2011TransformingAutoEncoders,2017DynamicRouting,2018MatrixCapsules}, some capsule-based methods are proposed. Liu et al.~\cite{2019EmployingDeepPart} incorporated the part-object relationships in salient object detection, which is implemented by the Capsule Network. Qi et al.~\cite{2019MultiScaleCapsule} proposed CapSalNet, which includes a multi-scale capsule attention module and multi-crossed layer connections for salient object detection. Readers could refer to some surveys~\cite{2014SalientObjectSurvey,2019SalientObjectDectionSurvey} for comprehensive understandings of 2D salient object detection.

\paragraph{Discussions}

The 2D object localization without classification are widely used in robotic grasping tasks but in a junior level. During industrial scenarios, the mechanical components are usually with fixed shapes, and many of them could be localized through fitting shape primitives. In some other grasping scenarios, the background priors or color contract is utilized to obtain the salient object for grasping. In Dexnet 2.0~\cite{2017DexNet2}, the target objects are laid on a workspace with green color, and they are easily segmented using color background subtraction.

\subsubsection{3D localization without classification}
\label{sec:2-1-2}

\begin{figure}[htbp]
\centering
\includegraphics[scale=0.4]{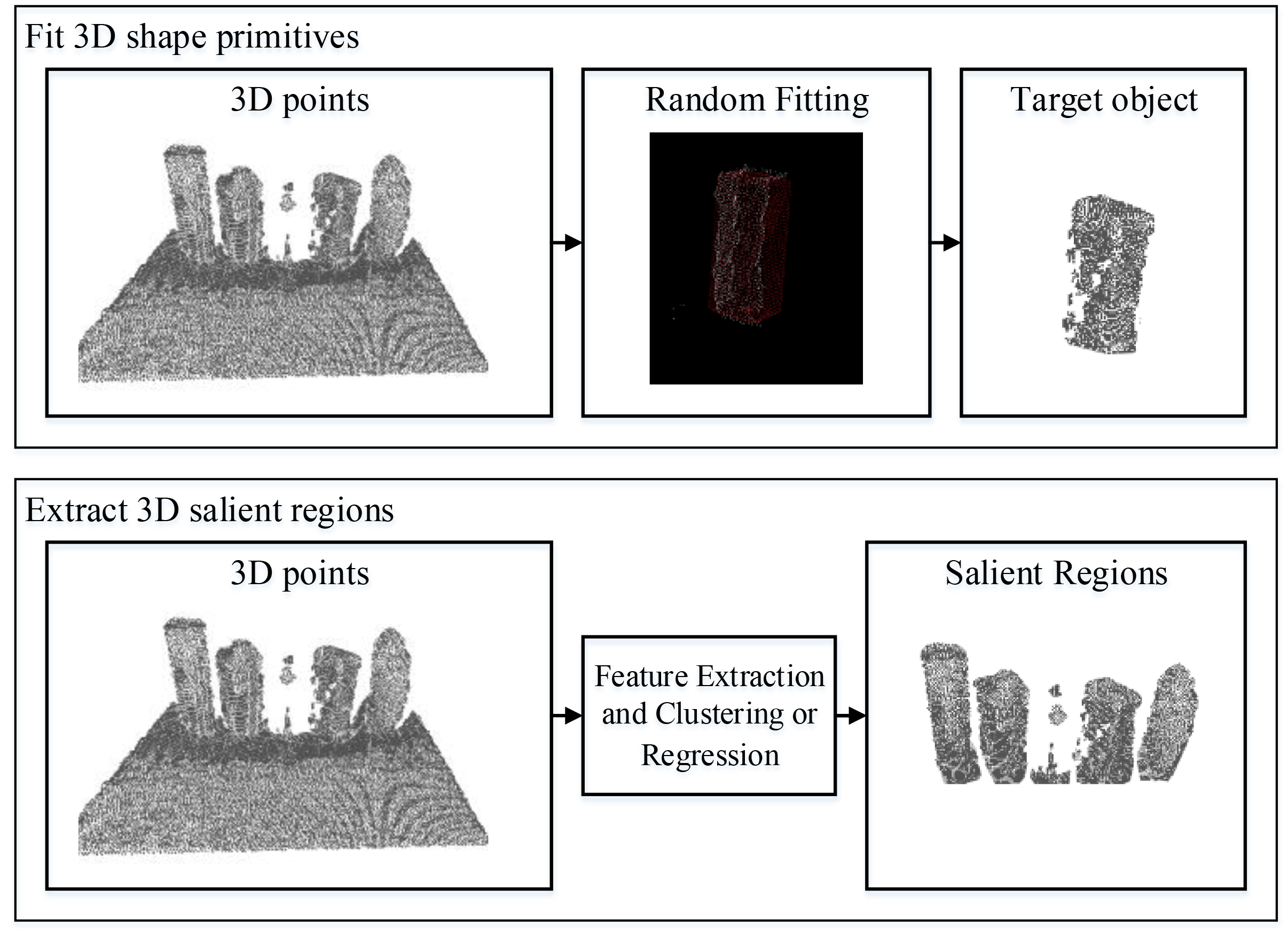}
\caption{Typical functional flow-chart of 3D object localization without classification.}
\label{fig:3dlocalization}
\end{figure}

This kind of methods deal with 3D point cloud inputs, which are usually partial point clouds reconstructed from single-view depth images in robotic grasping tasks. According to whether the object's 3D shape is known or not, methods can also be divided into methods of fitting 3D shape primitives and methods of salient 3D object detection. Typical functional flow-chart of 3D object localization without classification is illustrated in Fig.~\ref{fig:3dlocalization}.

\paragraph{Fitting 3D shape primitives}

The shape of the target object could be a sphere, a cylinder or a box, and these shapes could be regarded as 3D shape primitives. There exist lots of methods aiming at fitting 3D shape primitives, such as RANdom SAmple Consensus (RANSAC)~\cite{1981RandomANSAC}-based methods, Hough-like voting methods~\cite{2005EfficientHough} and other clustering techniques~\cite{2009CloseRangeScene,2012RobustlySegmenting}. These methods deal with different kinds of inputs and have been applied in areas like modeling, rendering and animation. Aiming at object localization and robotic grasping tasks, the input data is a partial point cloud, where the object is incomplete, and the ambition is to find the points that can constitute one of the 3D shape primitives. Some methods~\cite{2013LinearApproach,2015SeparatingObjects} detect planes at object boundaries and assemble them. Jiang et al.~\cite{2013LinearApproach} and Khan et al.~\cite{2015SeparatingObjects} explored the 3D structures in an indoor scene and estimated their geometry using cuboids. Rabbani et al.~\cite{2005EfficientHough} presented an efficient Hough transform for automatic detection of cylinders in point clouds. Some methods~\cite{2009CloseRangeScene,2012RobustlySegmenting} conduct primitive fitting after segmenting the scene. Rusu et al.~\cite{2009CloseRangeScene} used a combination of robust shape primitive models with triangular meshes to create a hybrid shape-surface representation optimal for robotic grasping. Goron et al.~\cite{2012RobustlySegmenting} presented a method to locate the best parameters for cylindrical and box-like objects in a cluttered scene. They increased the robustness of RANSAC fits when dealing with clutter through employing a set of inlier filters and the use of Hough voting. They provided robust results and models that are relevant for grasp estimation. Readers could refer to the survey~\cite{2019Survey3DPrimitives} for more details.

\paragraph{3D salient object detection}

Compared with 2D salient object detection, 3D salient object detection consumes many kinds of 3D data, such as depth image and point cloud. Although above 2D salient object detection methods have achieved superior performance, they still remain challenging in some complex scenarios, where depth information could provide much assistance. RGB-D saliency detection methods usually utilize hand-crafted or deep learning-based features from RGB-D images and fuse them in different ways. Peng et al.~\cite{2014RgbdSOD} proposed a simple fusion strategy which extends RGB-based saliency models by incorporating depth-induced saliency. Ren et al.~\cite{2015ExploitingGlobal} exploited the normalized depth prior and the global-context surface orientation prior for salient object detection. Qu et al.~\cite{2017RgbdSalientOD} trained a CNN-based model which fuses different low level saliency cues into hierarchical features for detecting salient objects in RGB-D images. Chen et al.~\cite{2019MultiModelFusionSOD,2019CnnRGBDSOD} utilized two-stream CNNs-based models with different fusion structures. Chen and Li~\cite{2018ProgressivelyComplementarity} further proposed a progressively complementarity-aware fusion network for RGB-D salient object detection, which is more effective than early-fusion methods~\cite{2017DeeplySupervisedSOD} and late-fusion methods~\cite{2018AdvancedDLSOD}. Piao et al.~\cite{2019DepthInduced} proposed a depth-induced multi-scale recurrent attention network (DMRANet) for saliency detection, which achieves dramatic performance especially in complex scenarios. Pang et al.~\cite{2020HierarchicalDynamicFN} proposed a hierarchical dynamic filtering network (HDFNet) and a hybrid enhanced loss. Li et al.~\cite{2020CMWNet} proposed a Cross-Modal Weighting (CMW) strategy to encourage comprehensive interactions between RGB and depth channels. These methods demonstrate remarkable performance of RGB-D SOD.

Aiming at 3D point cloud input, lots of methods are proposed to detect saliency maps of a complete object model~\cite{2019PointcloudSaliency}, whereas, our ambitious is to locate the salient object from the 3D scene inputs. Kim et al.~\cite{2008SegmentationSalientRegions} described a segmentation method for extracting salient regions in outdoor scenes using both 3D point clouds and RGB image. Bhatia et al.~\cite{2013SegmentingSOD3D} proposed a top-down approach for extracting salient objects/regions in 3d point clouds of indoor scenes.They first segregates significant planar regions, and extracts isolated objects present in the residual point cloud. Each object is then ranked for saliency based on higher curvature complexity of the silhouette.

\paragraph{Discussions}

3D object localization is widely used in robotic grasping tasks but also in a junior level. In Rusu et al.~\cite{2009CloseRangeScene} and Goron et al.~\cite{2012RobustlySegmenting}, fitting 3D shape primitives has been successfully applied into robotic grasping tasks. In Zapata-Impata et al.~\cite{2019FastGeometry}, the background is first filtered out using the height constraint, and the table is filtered out by fitting a plane using RANSAC~\cite{1981RandomANSAC}. The remained point cloud is clustered and $K$ object's clouds are achieved finally. There also exist some other ways to remove the background points through fitting background points using existing full 3D point cloud. These methods are successfully applied into robotic grasping tasks.

\subsection{Object detection}
\label{sec:2-2}

The task of object detection is to detect instances of objects of a certain class, which can be treated as a localization task plus a classification task. Usually, the shapes of the target objects are unknown, and accurate salient regions are hardly achieved. Therefore, the regularly bounding boxes are used for general object localization and classification tasks, and the outputs of object detection are bounding boxes with class labels. Based on whether using region proposals or not, the methods can be divided into two-stage methods and one-stage methods. These methods are summarized respectively in Table~\ref{tab:detection} aiming at 2D or 3D inputs.

\begin{table*}[htbp]
\caption{Methods of object detection.}
    \begin{tabular}{ m{1.5cm} m{8.0cm} m{6.5cm} }
    \hline
    Methods & Two-stage methods & One-stage methods \\
    \hline
    2D detection & SIFT~\cite{1999ObjectRecognition}, FAST~\cite{2005FAST}, SURF~\cite{2006SURF}, ORB~\cite{2011ORB}, OverFeat~\cite{2013OverFeat}, Erhan et al.~\cite{2014ScalableObjectDetection}, Szegedy et al.~\cite{2014ScalableHighQuality}, RCNN~\cite{2014RCNN}, Fast R-CNN~\cite{2015FastRCNN}, Faster RCNN~\cite{2015FasterRCNN}, R-FCN~\cite{2016RFCN}, FPN~\cite{2017FeaturePyramidNetworks} & YOLO~\cite{2016YOLO}, SSD~\cite{2016SSD}, YOLOv2~\cite{2017Yolo9000}, RetinaNet~\cite{2017RetinaNet}, YOLOv3~\cite{2018YOLOv3}, FCOS~\cite{2019FCOS}, CornerNet~\cite{2018CornerNet}, ExtremeNet~\cite{2019ExtremeNet}, CenterNet~\cite{2019CenterNetO,2019CenternetK}, CentripetalNet~\cite{2020Centripetalnet}, YOLOv4~\cite{2020YOLOv4} \\
    \hline
    3D detection & Spin Images~\cite{1997SpinImages}, 3D Shape Context~\cite{20043DSC}, FPFH~\cite{2009FPFH}, CVFH~\cite{2011CadRecognition6DOF}, SHOT~\cite{2014SHOT}, Sliding Shapes~\cite{2014SlidingShapes}, Frustum PointNets~\cite{2018FrustumPointnets}, PointFusion~\cite{2018PointFusion}, FrustumConvNet~\cite{2019FrustumConvNet}, Deep Sliding Shapes~\cite{2016DeepSlidingshapes}, MV3D~\cite{2017MV3D}, MMF~\cite{2019MultiMultiFusion}, Part-A$^{2}$~\cite{2020PartA2Net},  PV-RCNN~\cite{2020PV-RCNN}, PointRCNN~\cite{2019PointRCNN}, STD~\cite{2019STD3DDetection}, VoteNet~\cite{2019VoteNet}, MLCVNet~\cite{2020MLCVNet}, H3DNet~\cite{2020H3DNet}, ImVoteNet~\cite{2020IMVoteNet} & VoxelNet~\cite{2018VoxelNet}, SECOND~\cite{2018Second}, PointPillars~\cite{2019PointPillars}, TANet~\cite{2020TANet}, HVNet~\cite{2020HVNet3D}, 3DSSD~\cite{20203DSSD}, Point-GNN~\cite{2020PointGNN3D}, DOPS~\cite{2020DOPS}, Associate-3Ddet~\cite{2020Associate3Ddet} \\
    \hline
    \end{tabular}
\label{tab:detection}
\end{table*}

\subsubsection{2D object detection}
\label{sec:2-2-1}

\begin{figure}[htbp]
\centering
\includegraphics[scale=0.4]{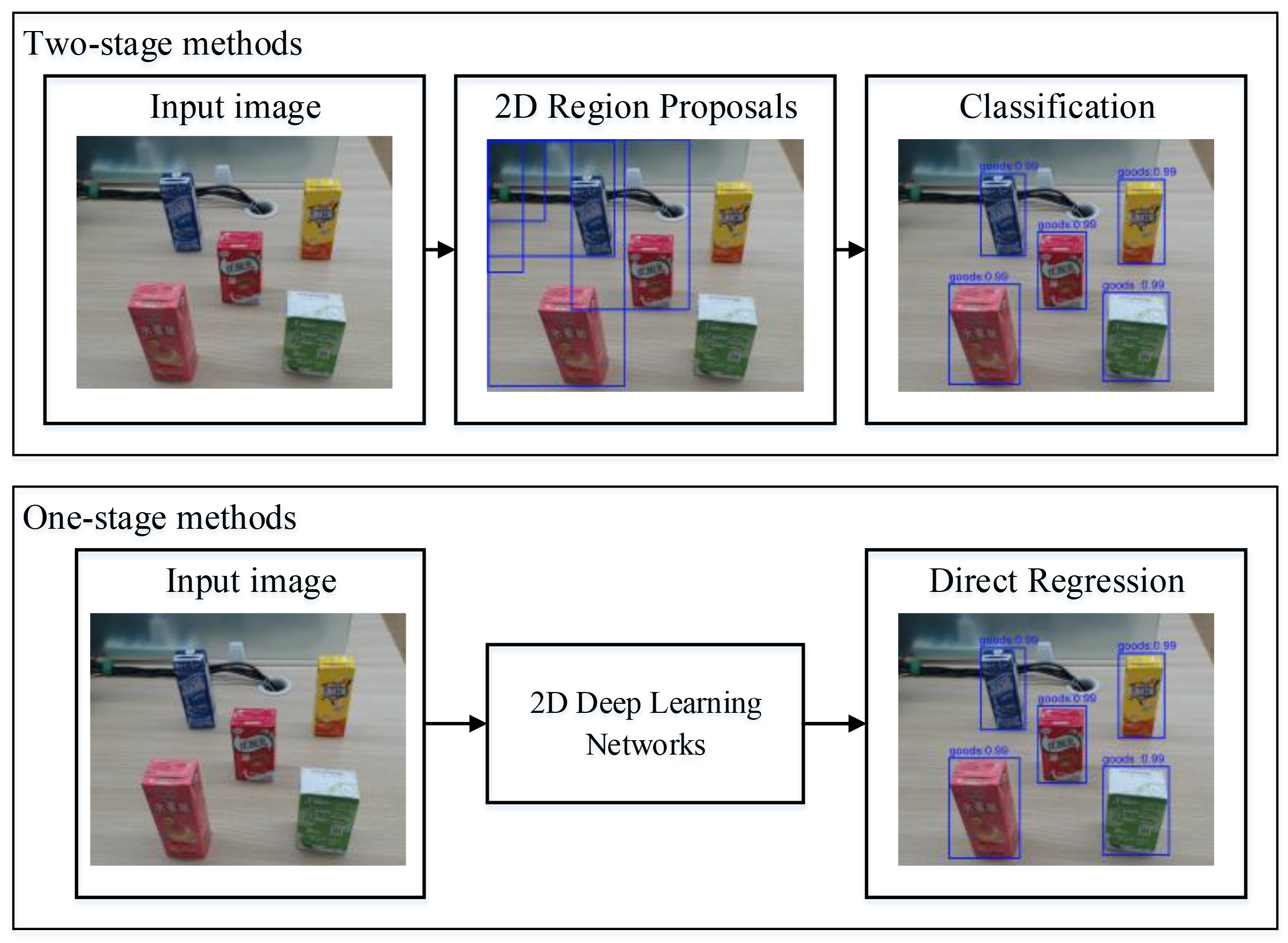}
\caption{Typical functional flow-chart of 2D object detection.}
\label{fig:2ddetection}
\end{figure}

2D object detection means detecting the target objects in 2D images by computing their 2D bounding boxes and categories. The most popular way of 2D detection is to generate object proposals and conduct classification, which is the two-stage methods. With the development of deep learning networks, especially Convolutional Neural Network (CNN), two-stage methods are improved extremely. In addition, large number of one-stage methods are proposed which achieved high accuracies with high speed. Typical functional flow-chart of 2D object detection is illustrated in Fig.~\ref{fig:2ddetection}.

\paragraph{Two-stage methods}

The two-stage methods can be referred as region proposal-based methods. Most of the traditional methods utilize the sliding window strategy to obtain the bounding boxes first, and then utilize feature descriptions of the bounding boxes for classification. Large number of hand-crafted global descriptors and local descriptors are proposed, such as SIFT~\cite{1999ObjectRecognition}, FAST~\cite{2005FAST}, SURF~\cite{2006SURF}, ORB~\cite{2011ORB}, and so on. Based on these descriptors, researchers trained classifiers, such as neural networks, Support Vector Machine (SVM) or Adaboost, to conduct 2D detection. There exist some disadvantages of traditional detection methods. For example, the sliding windows should be predefined for specific objects, and the hand-crafted features are not representative enough for a strong classifier.

With the development of deep learning, region proposals could be computed with a deep neural network. OverFeat~\cite{2013OverFeat} trained a fully connected layer to predict the box coordinates for the localization task that assumes a single object. Erhan et al.~\cite{2014ScalableObjectDetection} and Szegedy et al.~\cite{2014ScalableHighQuality} generated region proposals from a network whose last fully connected layer simultaneously predicts multiple boxes. Besides, deep neural networks extract more representative features than hand-crafted features, and training classifiers using CNN~\cite{2012DCNN} features highly improved the performance. R-CNN~\cite{2014RCNN} uses Selective Search (SS)~\cite{2013SelectiveSearch} methods to generate region proposals, uses CNN to extract features and trains classifiers using SVM. This traditional classifier is replaced by directly regressing the bounding boxes using the Region of Interest (ROI) feature vector in Fast R-CNN~\cite{2015FastRCNN}. Faster R-CNN~\cite{2015FasterRCNN} is further proposed by replacing SS with the Region Proposal Network (RPN), which is a kind of fully convolutional network (FCN)~\cite{2015FullyCN} and can be trained end-to-end specifically for the task of generating detection proposals. This design is also adopted in other two-stage methods, such as R-FCN~\cite{2016RFCN}, FPN~\cite{2017FeaturePyramidNetworks}. Generally, two-stage methods achieve a higher accuracy, whereas need more computing resources or computing time.

\paragraph{One-stage methods}

The one-stage methods can also be referred as regression-based methods. Compared to two-stage approaches, the single-stage pipeline skips separate object proposal generation and predicts bounding boxes and class scores in one evaluation. YOLO~\cite{2016YOLO} conducts joint grid regression, which simultaneously predicts multiple bounding boxes and class probabilities for those boxes. YOLO is not suitable for small objects, since it only regress two bounding boxes for each grid. SSD~\cite{2016SSD} predicts category scores and box offsets for a fixed set of anchor boxes produced by the sliding window. Compared with YOLO, SSD is faster and much more accurate. YOLOv2~\cite{2017Yolo9000} also adopts sliding window anchors for classification and spatial location prediction so as to achieve a higher recall than YOLO. RetinaNet~\cite{2017RetinaNet} proposed the focal loss function by reshaping the standard cross entropy loss so that detector will put more focus on hard, misclassified examples during training. RetinaNet achieved comparable accuracy of two-stage detectors with high detection speed. Compare with YOLOv2, YOLOv3~\cite{2018YOLOv3} and YOLOv4~\cite{2020YOLOv4} are further improved with a bunch of improvements, which shows large performance improvements without sacrificing the speed, and is more robust in dealing with small objects. There also exist some anchor-free methods, which doesn't utilize the anchor bounding boxes, such as FCOS~\cite{2019FCOS}, CornerNet~\cite{2018CornerNet}, ExtremeNet~\cite{2019ExtremeNet}, CenterNet~\cite{2019CenterNetO,2019CenternetK} and CentripetalNet~\cite{2020Centripetalnet}. Further reviews of these works can refer to recent surveys~\cite{2019ObjectDetSurvey,2019ObjectDetReview,2020DLObjDetSurvey,2020ReviewObjectDetection}.

\paragraph{Discussions}

The 2D object detection methods are widely used in 2D planar robotic grasping tasks. This part can refer to Section~\ref{4-1-2}.

\subsubsection{3D object detection}
\label{sec:2-2-2}

\begin{figure}[htbp]
\centering
\includegraphics[scale=0.3]{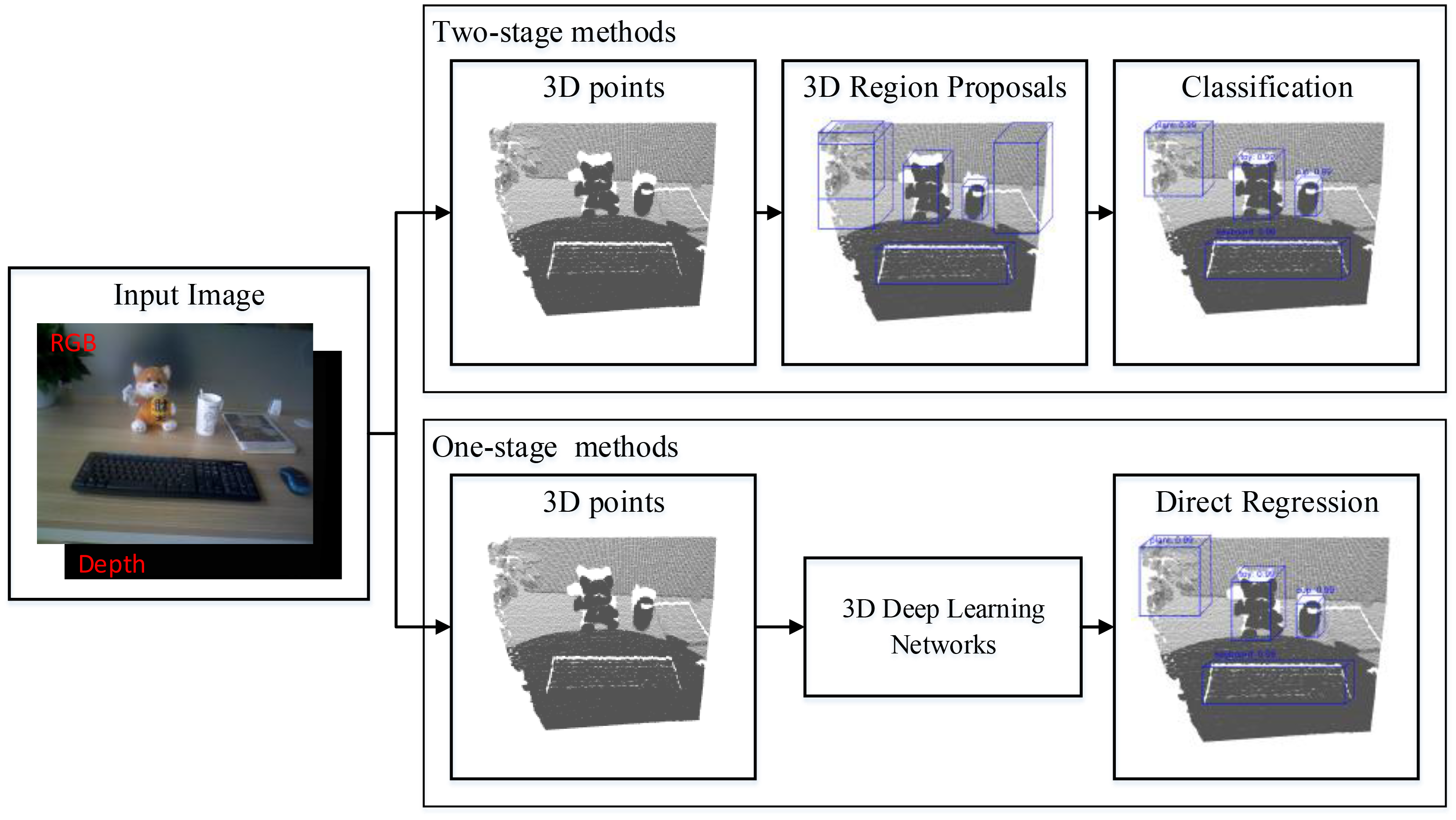}
\caption{Typical functional flow-chart of 3D object detection.}
\label{fig:3ddetection}
\end{figure}

3D object detection aims at finding the amodel 3D bounding box of the target object, which means finding the 3D bounding box that a complete target object occupies. 3D object detection is deeply explored in outdoor scenes and indoor scenes. Aiming at robotic grasping tasks, we can obtain the 2D and 3D information of the scene through RGB-D data, and general 3D object detection methods could be used. Similar with 2D object detection tasks, two-stage methods and one-stage methods both exist. The two-stage methods refer to region proposal-based methods and one-stage methods refer to regression-based methods. Typical functional flow-chart of 3D object detection is illustrated in Fig.~\ref{fig:3ddetection}.

\paragraph{Two-stage methods}

Traditional 3D detection methods usually aim at objects with known shapes. The 3D object detection problem is transformed into a detection and 6D object pose estimation problem. Many hand-crafted 3D shape descriptors, such as Spin Images~\cite{1997SpinImages}, 3D Shape Context~\cite{20043DSC}, FPFH~\cite{2009FPFH}, CVFH~\cite{2011CadRecognition6DOF}, SHOT~\cite{2014SHOT}, are proposed, which can locate the object proposals. In addition, the accurate 6D pose of the target object could be achieved through local registration. This part is introduced in Section~\ref{sec:3-1-2}. However, these methods face difficulties in general 3D object detection tasks. Aiming at general 3D object detection tasks, the 3D region proposals are widely used. Traditional methods train classifiers, such as SVM, based on the 3D shape descriptors. Sliding Shapes~\cite{2014SlidingShapes} is proposed which slides a 3D detection window in 3D space and extract features from the 3D point cloud to train an Exemplar-SVM classifier~\cite{2011ExemplarSvm}. With the development of deep learning, the 3D region proposals could be generated efficiently, and the 3D bounding boxes could be regressed using features from deep neural networks rather than training traditional classifiers. There exist various methods of generating 3D object proposals, which can be roughly divided into three kinds, which are frustum-based methods~\cite{2018FrustumPointnets,2018PointFusion,2019FrustumConvNet}, global regression-based methods~\cite{2016DeepSlidingshapes,2017MV3D,2019MultiMultiFusion} and local regression-based methods.

Frustum-based methods generate object proposals using mature 2D object detectors, which is a straightforward way. Frustum PointNets~\cite{2018FrustumPointnets} leverages a 2D CNN object detector to obtain 2D regions, and the lifted frustum-like 3D point clouds become 3D region proposals. The amodel 3D bounding boxes are regressed from features of the segmented points within the proposals based on PointNet~\cite{2017Pointnet}. PointFusion~\cite{2018PointFusion} utilized Faster R-CNN~\cite{2015FasterRCNN} to obtain the image crop first, and deep features from the corresponding image and the raw point cloud are densely fused to regress the 3D bounding boxes. FrustumConvNet~\cite{2019FrustumConvNet} also utilizes the 3D region proposals lifted from the 2D region proposal and generates a sequence of frustums for each region proposal.

Global regression-based methods generate 3D region proposals from feature representations extracted from single or multiple inputs. Deep Sliding Shapes~\cite{2016DeepSlidingshapes} proposed the first 3D Region Proposal Network (RPN) using 3D convolutional neural networks (ConvNets) and the first joint Object Recognition Network (ORN) to extract geometric features in 3D and color features in 2D to regress 3D bounding boxes. MV3D~\cite{2017MV3D} represents the point cloud using the bird's-eye view and employs 2D convolutions to generate 3D proposals. The region-wise features obtained via ROI pooling for multi-view data are fused to jointly predict the 3D bounding boxes. MMF~\cite{2019MultiMultiFusion} proposed a multi-task multi-sensor fusion model for 2D and 3D object detection, which generates a small number of high-quality 3D detections using multi-sensor fused features, and applies ROI feature fusion to regress more accurate 2D and 3D boxes. Part-A$^{2}$~\cite{2020PartA2Net} predicts intra-object part locations and generates 3D proposals by feeding the point cloud to an encoder-decoder network. A RoI-aware point cloud pooling is proposed to aggregate the part information from each 3D proposal, and a part-aggregation network is proposed to refine the results. PV-RCNN~\cite{2020PV-RCNN} utilized voxel CNN with 3D sparse convolution~\cite{2017SubmanifoldSparseConv,20183DSparseConv} for feature encoding and proposals generation, and proposed a voxel-to-keypoint scene encoding via voxel set abstraction and a keypoint-to-grid RoI feature abstraction for proposal refinement. PV-RCNN achieved remarkable 3D detection performance on outdoor scene datasets.

Local regression-based methods mean generating point-wise 3D region proposals. PointRCNN~\cite{2019PointRCNN} extracts point-wise feature vectors from the input point cloud and generates 3D proposal from each foreground point computed through segmentation. Point cloud region pooling and canonical 3D bounding box refinement are then conducted. STD~\cite{2019STD3DDetection} designs spherical anchors and a strategy in assigning labels to anchors to generate accurate point-based proposals, and a PointsPool layer is proposed to generate dense proposal features for the final box prediction. VoteNet~\cite{2019VoteNet} proposed a deep hough voting strategy to generate 3D vote points from sampled 3D seeds points. The 3D vote points are clustered to obtain object proposals which will be further refined. MLCVNet~\cite{2020MLCVNet} proposed Multi-level Context VoteNet which considers the contextual information between the objects. H3DNet~\cite{2020H3DNet} predicts a hybrid set of geometric primitives such as centers, face centers and edge centers of the 3d bounding boxes, and formulates 3D object detection as regressing and aggregating these geometric primitives. A matching and refinement module is then utilized to classify object proposals and fine-tune the results. Compared with point cloud input-only VoteNet~\cite{2019VoteNet}, ImVoteNet~\cite{2020IMVoteNet} additionally extracts geometric and semantic features from the 2D images, and fuses the 2D features into the 3D detection pipeline, which achieved remarkable 3D detection performance on indoor scene datasets.

\paragraph{One-stage methods}

One-stage methods directly predict class probabilities and regress the 3D amodal bounding boxes of the objects using a single-stage network. These methods do not need region proposal generation or post-processing. VoxelNet~\cite{2018VoxelNet} divides a point cloud into equally spaced 3D voxels and transforms a group of points within each voxel into a unified feature representation. Through convolutional middle layers and the region proposal network, the final results are obtained. Compared with VoxelNet, SECOND~\cite{2018Second} applies sparse convolution layers~\cite{20183DSparseConv} for parsing the compact voxel features. PointPillars~\cite{2019PointPillars} converts a point cloud to a sparse pseudo-image, which is processed into a high-level representation through a 2D convolutional backbone. The features from the backbone are used by the detection head to predict 3D bounding boxes for objects. TANet~\cite{2020TANet} proposed a Triple Attention (TA) module and a Coarse-to-Fine Regression (CFR) module, which focuses on the 3D detection of hard objects and the robustness to noisy points. HVNet~\cite{2020HVNet3D} proposed a hybrid voxel network which fuses voxel feature encoder (VFE) of different scales at point-wise level and projects into multiple pseudo-image feature maps. Above methods are mainly voxel-based 3D single stage detectors, and Yang et al.~\cite{20203DSSD} proposed a point-based 3D single stage object detector called 3DSSD, which contain a fusion sampling strategy in the downsampling process, a candidate generation layer, and an anchor-free regression head with a 3D center-ness assignment strategy. They achieved a good balance between accuracy and efficiency.
Point-GNN~\cite{2020PointGNN3D} utilized graph neural network on the point cloud and designed a graph neural network with an auto-registration mechanism which detects multiple objects in a single shot. DOPS~\cite{2020DOPS} proposed an object detection pipeline which utilizes a 3D sparse U-Net~\cite{2017SubmanifoldSparseConv} and a graph convolution module. Their method can jointly predict the 3D shapes of the objects. Associate-3Ddet~\cite{2020Associate3Ddet} learns to associate feature extracted from the real scene with more discriminative feature from class-wise conceptual models. Comprehensive review about 3D object detection could refer to the survey~\cite{20193DDeepLearningSurvey}.

\paragraph{Discussions}

3D object detection only presents the general shape of the target object, which is not sufficient to conduct a robotic grasp, and it is mostly used in autonomous driving areas. However, the estimated 3D bounding boxes could provide approximate grasp positions and provide valuable information for the collision detection.

\subsection{Object instance segmentation}
\label{sec:2-3}

Object instance segmentation refers to detecting the pixel-level or point-level instance objects of a certain class, which is closely related to object detection and semantic segmentation tasks. Two kinds of methods also exist, which are two-stage methods and one-stage methods. The two-stage methods refer to region proposal-based methods and one-stage methods refer to regression-based methods. The representative works of the two methods are shown in Table~\ref{tab:segmentation} aiming at 2D inputs and 3D inputs.

\begin{table*}[htbp]
\caption{Methods of object instance segmentation.}
    \begin{tabular}{ m{2.0cm} m{5.5cm} m{8.5cm} }
    \hline
    Methods & Two-stage methods & One-stage methods \\
    \hline
    2D instance segmentation & SDS~\cite{2014SimultaneousDaS}, MNC~\cite{2016MNCInstanceAwareSemantic}, PANet~\cite{2018PANet}, Mask R-CNN~\cite{2017MaskRCNN}, MaskLab~\cite{2018MaskLab}, HTC~\cite{2019HybridTaskCascade}, PointRend~\cite{2019PointRend}, FGN~\cite{2020FGNInstanceSeg} & DeepMask~\cite{2015DeepMask}, SharpMask~\cite{2016SharpMask}, InstanceFCN~\cite{2016InstanceFCN}, FCIS~\cite{2017FCIS}, TensorMask~\cite{2019TensorMask}, YOLACT~\cite{2019Yolact}, YOLACT++~\cite{2019Yolact++}, PolarMask~\cite{2019PolarMask}, SOLO~\cite{2019SOLO}, CenterMask~\cite{2020CenterMask}, BlendMask~\cite{2020BlendMask} \\
    \hline
    3D instance segmentation & GSPN~\cite{2019GSPN}, 3D-SIS~\cite{20193DSIS}, 3D-MPA~\cite{20203DMPA} & SGPN~\cite{2018SGPN}, MASC~\cite{2019MASC}, ASIS~\cite{2019ASIS}, JSIS3D~\cite{2019JSIS3D}, JSNet~\cite{2020JSNet}, 3D-BoNet~\cite{20193DBoNet}, LiDARSeg~\cite{2020LiDARSeg}, OccuSeg~\cite{2020OccuSeg} \\
    \hline
    \end{tabular}
\label{tab:segmentation}
\end{table*}

\subsubsection{2D object instance segmentation}
\label{sec:2-3-1}

\begin{figure}[htbp]
\centering
\includegraphics[scale=0.35]{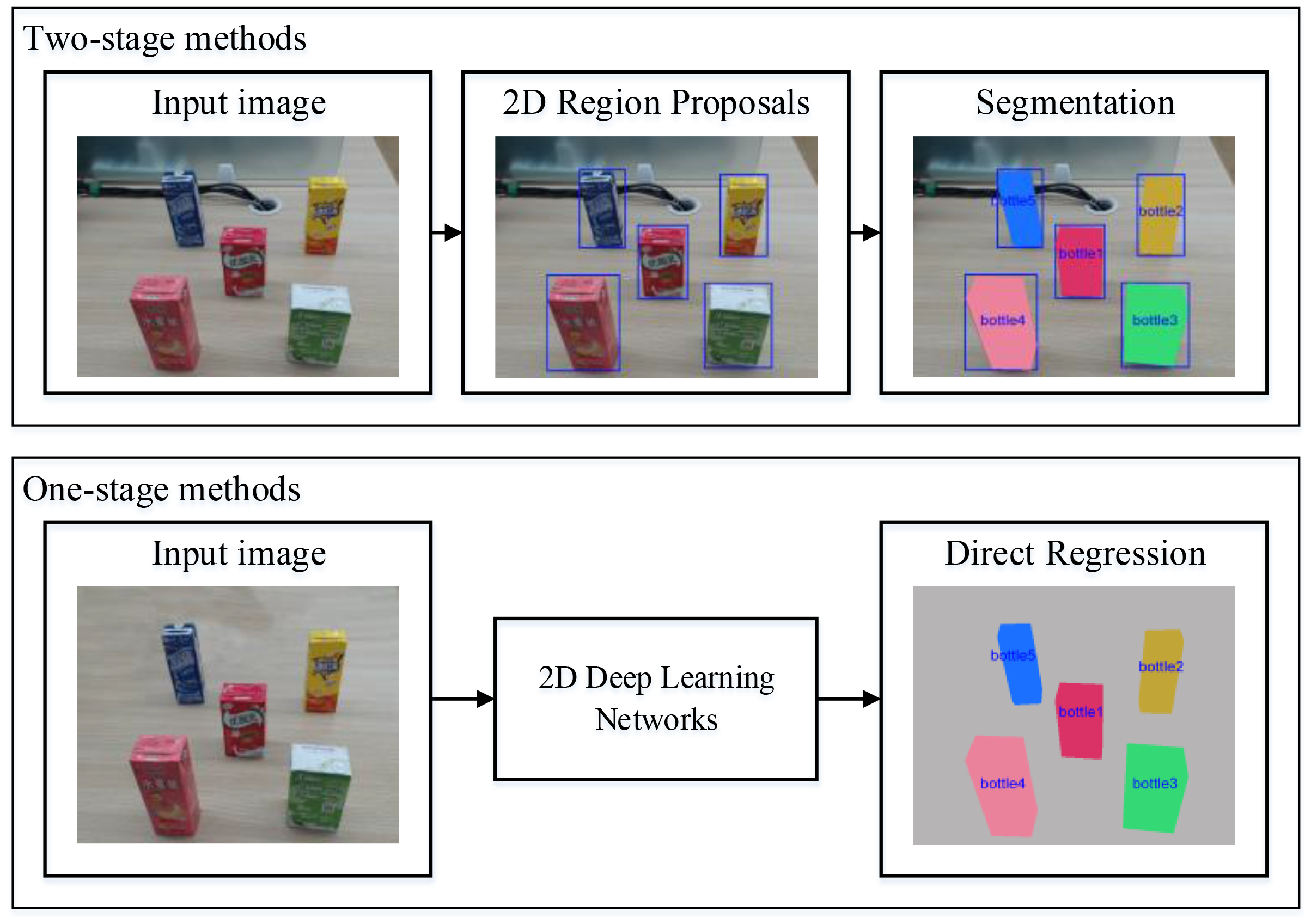}
\caption{Typical functional flow-chart of 2D object instance segmentation.}
\label{fig:2dsegmentation}
\end{figure}

2D object instance segmentation means detecting the pixel-level instance objects of a certain class from an input image, which is usually represented as masks. Two-stage methods follow the mature object detection frameworks, while one-stage methods conduct regression from the whole input image directly. Typical functional flow-chart of 2D object instance segmentation is illustrated in Fig.~\ref{fig:2dsegmentation}.

\paragraph{Two-stage methods}

This kind of methods could also be referred as region proposal-based methods. The mature 2D object detectors are used to generate bounding boxes or region proposals, and the object masks are then predicted within the bounding boxes. Lots of methods are based on convolutional neural networks (CNN). SDS~\cite{2014SimultaneousDaS} uses CNN to classify category-independent region proposals. MNC~\cite{2016MNCInstanceAwareSemantic} conducts instance segmentation via three networks, respectively differentiating instances, estimating masks, and categorizing objects. Path Aggregation Network (PANet)~\cite{2018PANet} was proposed which boosts the information flow in the proposal-based instance segmentation framework. Mask R-CNN~\cite{2017MaskRCNN} extends Faster R-CNN~\cite{2015FasterRCNN} by adding a branch for predicting an object mask in parallel with the existing branch for bounding box recognition, which achieved promising results. MaskLab~\cite{2018MaskLab} also builds on top of Faster R-CNN~\cite{2015FasterRCNN} and additionally produces semantic and instance center direction outputs. Chen et al.~\cite{2019HybridTaskCascade} proposed a framework called Hybrid Task Cascade (HTC), which performs cascaded refinement on object detection and segmentation jointly and adopts a fully convolutional branch to provide spatial context. PointRend~\cite{2019PointRend} performs point-based segmentation predictions at adaptively selected locations based on an iterative subdivision algorithm. PointRend can be flexibly applied to instance segmentation tasks by building on top of them, and yields significantly more detailed results. FGN~\cite{2020FGNInstanceSeg} proposed a Fully Guided Network (FGN) for few-shot instance segmentation, which introduces different guidance mechanisms into the various key components in Mask R-CNN~\cite{2017MaskRCNN}.

\paragraph{Single-stage methods}

This kind of methods could also be referred as regression-based methods, where the segmentation masks are predicted as well the objectness score. DeepMask~\cite{2015DeepMask}, SharpMask~\cite{2016SharpMask} and InstanceFCN~\cite{2016InstanceFCN} predict segmentation masks for the the object located at the center. FCIS~\cite{2017FCIS} was proposed as the fully convolutional instance-aware semantic segmentation method, where position-sensitive inside/outside score maps are used to perform object segmentation and detection. TensorMask~\cite{2019TensorMask} uses structured 4D tensors to represent masks over a spatial domain and presents a framework to predict dense masks. YOLACT~\cite{2019Yolact} breaks instance segmentation into two parallel subtasks, which are generating a set of prototype masks and predicting per-instance mask coefficients. YOLACT is the first real-time one-stage instance segmentation method and is improved by YOLACT++~\cite{2019Yolact++}. PolarMask~\cite{2019PolarMask} formulates the instance segmentation problem as predicting contour of instance through instance center classification and dense distance regression in a polar coordinate. SOLO~\cite{2019SOLO} introduces the notion of instance categories, which assigns categories to each pixel within an instance according to the instance's location and size, and converts instance mask segmentation into a classification-solvable problem. CenterMask~\cite{2020CenterMask} adds a novel spatial attention-guided mask (SAG-Mask) branch to anchor-free one stage object detector (FCOS~\cite{2019FCOS}) in the same vein with Mask R-CNN~\cite{2017MaskRCNN}. BlendMask~\cite{2020BlendMask} also builds upon the FCOS~\cite{2019FCOS} object detector, which uses a blender module to effectively predict dense per-pixel position-sensitive instance features and learn attention maps for each instance. Detailed reviews refer to the survey~\cite{2020EvolutionSegmentationSurvey,2020SurveyInstanceSegmentation}.

\paragraph{Discussions}

2D object instance segmentation is widely used in robotic grasping tasks. For example, SegICP~\cite{2017Segicp} utilize RGB-based object segmentation to obtain the points belong to the target objects. Xie et al.~\cite{2019TheBestOfBoth2DSeg} separately leverage RGB and Depth for unseen object instance segmentation. Danielczuk et al.~\cite{2019SegmentingUnknown3D} segments unknown 3d objects from real depth images using Mask R-CNN~\cite{2017MaskRCNN} trained on synthetic data.

\subsubsection{3D object instance segmentation}
\label{sec:2-3-2}

\begin{figure}[htbp]
\centering
\includegraphics[scale=0.3]{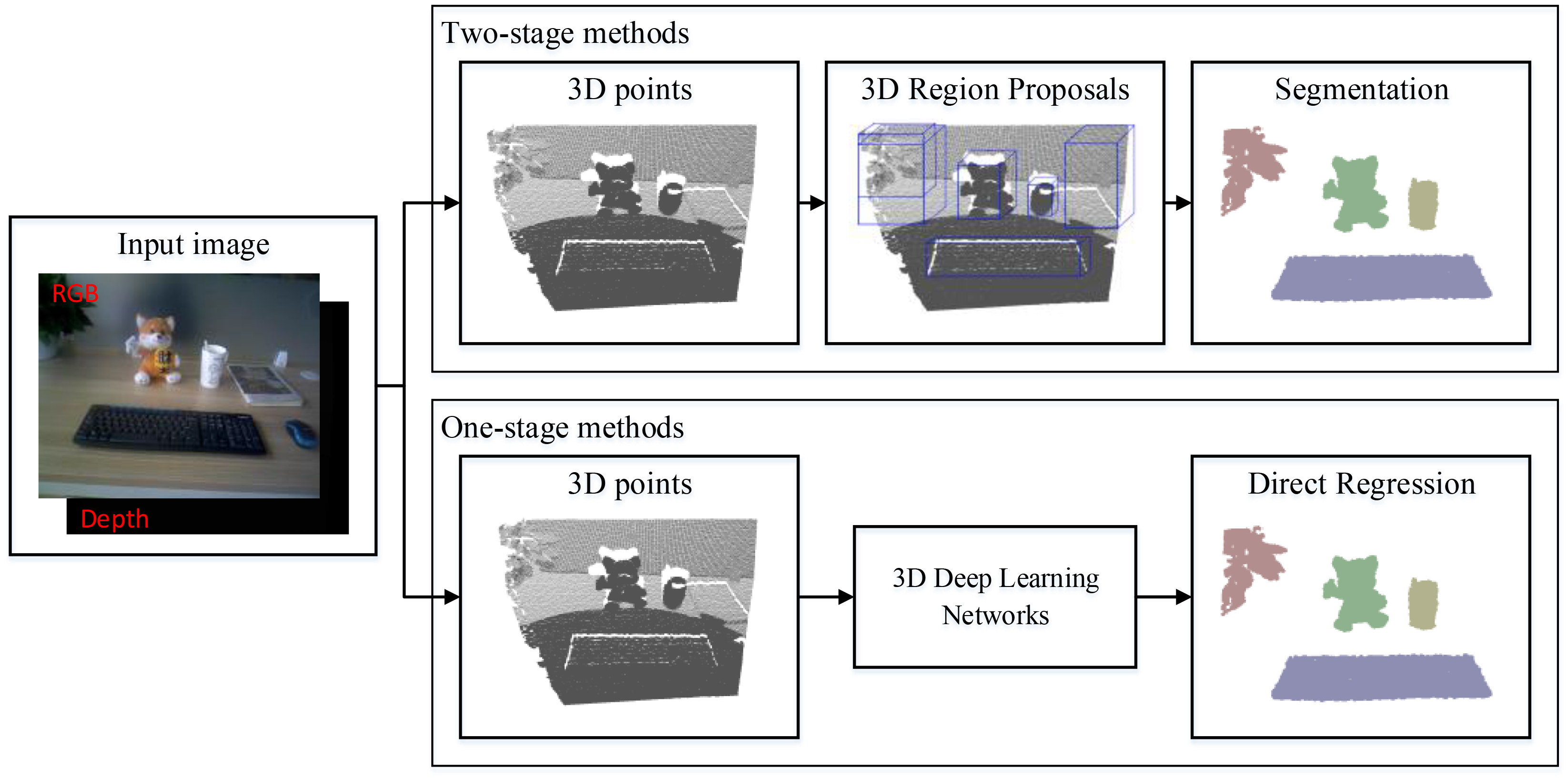}
\caption{Typical functional flow-chart of 3D object instance segmentation.}
\label{fig:3dsegmentation}
\end{figure}

3D object instance segmentation means detecting the point-level instance objects of a certain class from an input 3D point cloud. Similar to 2D object instance segmentation, two-stage methods need region proposals, while one-stage methods are proposal-free. Typical functional flow-chart of 3D object instance segmentation is illustrated in Fig.~\ref{fig:3dsegmentation}.

\paragraph{Two-stage methods}

This kind of methods could also be referred as proposal-based methods. General methods utilize the 2D or 3D detection results and conduct foreground or background segmentation in the corresponding frustum or bounding boxes. GSPN~\cite{2019GSPN} proposed the Generative Shape Proposal Network (GSPN) to generates 3D object proposals and the Region-PointNet framework to conduct 3D object instance segmentation. 3D-SIS~\cite{20193DSIS} leverages joint 2D and 3D end-to-end feature learning on both geometry and RGB input for 3D object bounding box detection and semantic instance segmentation. 3D-MPA~\cite{20203DMPA} predicts dense object centers based on learned semantic features from a sparse volumetric backbone, employes a graph convolutional network to explicitly model higher-order interactions between neighboring proposal features, and utilizes a multi proposal aggregation strategy other than NMS to obtain the final results.

\paragraph{Single-stage methods}

This kind of methods could also be referred as regression-based methods. Lots of methods learn to group per-point features to segment 3D instances. SGPN~\cite{2018SGPN} proposed the Similarity Group Proposal Network (SGPN) to predict point grouping proposals and a corresponding semantic class for each proposal, from which we can directly extract instance segmentation results. MASC~\cite{2019MASC} utilizes the sub-manifold sparse convolutions~\cite{2017SubmanifoldSparseConv,20183DSparseConv} to predict semantic scores for each point as well as the affinity between neighboring voxels at different scales. The points are then grouped into instances based on the predicted affinity and the mesh topology. ASIS~\cite{2019ASIS} learns semantic-aware point-level instance embedding and semantic features of the points belonging to the same instance are fused together to make per-point semantic predictions. JSIS3D~\cite{2019JSIS3D} proposed a multi-task point-wise network (MT-PNet) that simultaneously predicts the object categories of 3D points and embeds these 3D points into high dimensional feature vectors that allow clustering the points into object instances. JSNet~\cite{2020JSNet} also proposed a joint instance and semantic segmentation (JISS) module and designed an efficient point cloud feature fusion (PCFF) module to generate more discriminative features. 3D-BoNet~\cite{20193DBoNet} was proposed to directly regress 3D bounding boxes for all instances in a point cloud, while simultaneously predicting a point-level mask for each instance. LiDARSeg~\cite{2020LiDARSeg} proposed a dense feature encoding technique, a solution for single-shot instance prediction and effective strategies for handling severe class imbalances. OccuSeg~\cite{2020OccuSeg} proposed an occupancy-aware 3D instance segmentation scheme, which predicts the number of occupied voxels for each instance. The occupancy signal guides the clustering stage of 3D instance segmentation and OccuSeg achieves remarkable performance.

\paragraph{Discussions}

3D object instance segmentation is quite important in robotic grasping tasks. However, current methods mainly leverage 2D instance segmentation methods to obtain the 3D point cloud of the target object, which utilizes the advantages of RGB-D images. Nowadays 3D object instance segmentation is still a fast developing area, and it will be widely used in the future if its performance and speed improve a lot.

\section{Object Pose Estimation}
\label{sec:3}

In some 2D planar grasps, the target objects are constrained in the 2D workspace and are not piled up, the 6D object pose can be represented as the 2D position and the in-plane rotation angle. This case is relatively simple and is addressed quite well based on matching 2D feature points or 2D contour curves. In other 2D planar grasp and 6DoF grasp scenarios, the 6D object pose is mostly needed, which helps a robot to get aware of the 3D position and 3D orientation of the target object. The 6D object pose transforms the object from the object coordinate into the camera coordinate. We mainly focus on 6D object pose estimation in this section and divide 6D object pose estimation into three kinds, which are correspondence-based, template-based and voting-based methods. During the review of each kind of methods, both traditional methods and deep learning-based methods are reviewed.

\subsection{Correspondence-based methods}
\label{sec:3-1}

Correspondence-based 6D object pose estimation involves methods of finding correspondences between the observed input data and the existing complete 3D object model. When we want to solve this problem based on the 2D RGB image, we need to find correspondences between 2D pixels and 3D points of the existing 3D model. The 6D object pose can thus be recovered through Perspective-n-Point (PnP) algorithms~\cite{2009EPNP}. When we want to solve this problem based on the 3D point cloud lifted from the depth image, we need to find correspondences of 3D points between the observed partial-view point cloud and the complete 3D model. The 6D object pose can thus recovered through least square methods. The methods of correspondence-based methods are summarized in Table~\ref{tab:6d-corres}.

\begin{table*}[htbp]
\caption{Summary of correspondence-based 6D object pose estimation methods.}
    \begin{tabular}{ m{2.0cm} m{2.0cm} m{5.0cm} m{7.0cm} }
    \hline
    Methods & Descriptions & Traditional methods & Deep learning-based methods \\
    \hline
    2D image-based methods & Find correspondences between 2D pixels and 3D points, and use PnP methods & SIFT~\cite{1999ObjectRecognition}, FAST~\cite{2005FAST}, SURF~\cite{2006SURF}, ORB~\cite{2011ORB} & LCD~\cite{2020LCD}, BB8~\cite{2017BB8}, Tekin et al.~\cite{2018RealTime}, Crivellaro et al.~\cite{2017Robust3D}, KeyPose~\cite{2020KeyPose}, Hu et al.~\cite{2020SingleStage2D3D}, HybridPose~\cite{2020HybridPose}, Hu et al.~\cite{2018SegmentationDriven}, DPOD~\cite{2019DPOD}, Pix2pose~\cite{2019Pix2pose}, EPOS~\cite{2020EPOS} \\
    \hline
    3D point cloud-based methods & Find correspondences between 3D points & Spin Images~\cite{1997SpinImages}, 3D Shape Context~\cite{20043DSC}, FPFH~\cite{2009FPFH}, CVFH~\cite{2011CadRecognition6DOF}, SHOT~\cite{2014SHOT} &  3DMatch~\cite{20173dmatch},  3DFeat-Net~\cite{20183dfeatnet}, Gojcic et al.~\cite{2019PerfectMatch}, Yuan et al.~\cite{2020SelfSupervisedPSL}, StickyPillars~\cite{2020Stickypillars} \\
    \hline
    \end{tabular}
\label{tab:6d-corres}
\end{table*}

\subsubsection{2D image-based methods}
\label{sec:3-1-1}

\begin{figure}[htbp]
\centering
\includegraphics[scale=0.34]{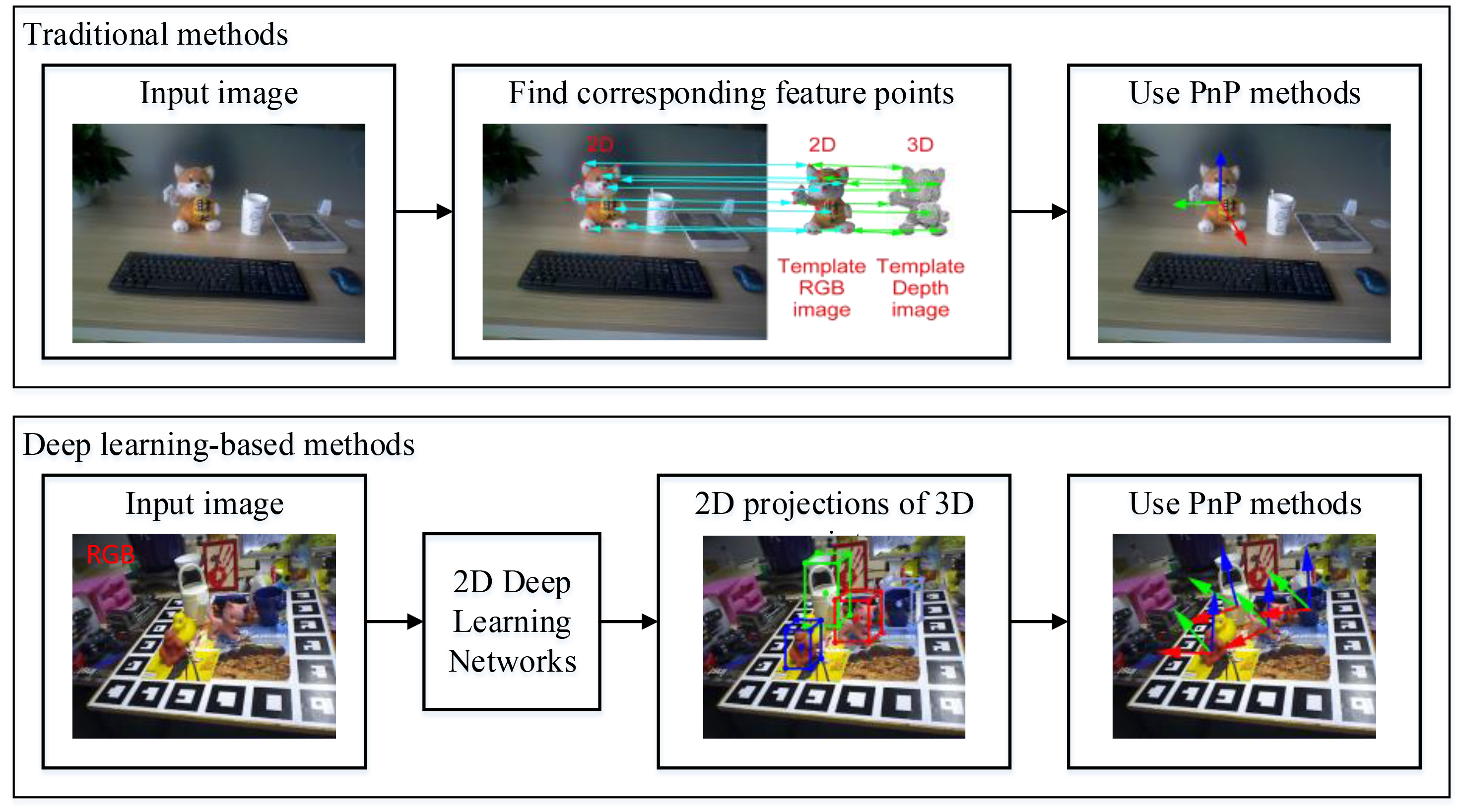}
\caption{Typical functional flow-chart of 2D correspondence-based 6D object pose estimation methods. Data from the lineMod dataset~\protect\cite{2012ModelBased}.}
\label{fig:2d-corres}
\end{figure}

When using the 2D RGB image, correspondence-based methods mainly target on the objects with rich texture through the matching of 2D feature points, as shown in Fig.~\ref{fig:2d-corres}. Multiple images are first rendered by projecting the existing 3D models from various angles and each object pixel in the rendered images corresponds to a 3D point. Through matching 2D feature points on the observed image and the rendered images~\cite{2004StableRealTime,2005Monocular}, the 2D-3D correspondences are established. Other than rendered images, the keyframes in keyframe-based SLAM approaches~\cite{2015ORBSLAM} could also provide 2D-3D correspondences for 2D keypoints. The common 2D descriptors such as SIFT~\cite{1999ObjectRecognition}, FAST~\cite{2005FAST}, SURF~\cite{2006SURF}, ORB~\cite{2011ORB}, etc., are usually utilized for the 2D feature matching. Based on the 2D-3D correspondences, the 6D object pose can be calculated with Perspective-n-Point (PnP) algorithms~\cite{2009EPNP}. However, these 2D feature-based methods fail when the objects do not have rich texture.

With the development of deep neural networks such as CNN, representative features could be extracted from the image. A straightforward way is to extract discriminative feature points~\cite{2016LIFT,2019Glampoints} and match them using the representative CNN features. Yi et al.~\cite{2016LIFT} presented a SIFT-like feature descriptor. Truong et al.~\cite{2019Glampoints} presented a method to greedily learn accurate match points. Superpoint~\cite{2018Superpoint} proposed a self-supervised framework for training interest point detectors and descriptors, which shows advantages over a few traditional feature detectors and descriptors. LCD~\cite{2020LCD} particularly learns a local cross-domain descriptor for 2D image and 3D point cloud matching, which contains a dual auto-encoder neural network that maps 2D and 3D inputs into a shared latent space representation.

There exists another kind of methods~\cite{2017BB8,2018RealTime,2017Robust3D,2018SegmentationDriven}, which uses the representative CNN features to predict the 2D locations of 3D points, as shown in Fig.~\ref{fig:2d-corres}. Since it's difficult to selected the 3D points to be projected, many methods utilize the eight vertices of the object's 3D bounding box. Rad and Lepetit~\cite{2017BB8} predicts 2D projections of the corners of their 3D bounding boxes and obtains the 2D-3D correspondences. Different with them, Tekin et al.~\cite{2018RealTime} proposed a single-shot deep CNN architecture that directly detects the 2D projections of the 3D bounding box vertices without posteriori refinements. Some other methods utilize feature points of the 3D object. Crivellaro et al.~\cite{2017Robust3D} predicts the pose of each part of the object in the form of the 2D projections of a few control points with the assistance of a Convolutional Neural Network (CNN). KeyPose~\cite{2020KeyPose} predicts object poses using 3D keypoints from stereo input, and is suitable for transparent objects. Hu et al.~\cite{2020SingleStage2D3D} further predicts the 6D object pose from a group of candidate 2D-3D correspondences using deep learning networks in a single-stage manner, instead of RANSAC-based Perspective-n-Point (PnP) algorithms. HybridPose~\cite{2020HybridPose} predicts a hybrid intermediate representation to express different geometric information in the input image, including keypoints, edge vectors, and symmetry correspondences. Some other methods predict 3D positions for all the pixels of the object. Hu et al.~\cite{2018SegmentationDriven} proposed a segmentation-driven 6D pose estimation framework where each visible part of the object contributes to a local pose prediction in the form of 2D keypoint locations. The pose candidates are them combined into a robust set of 2D-3D correspondences from which the reliable pose estimation result is computed. DPOD~\cite{2019DPOD} estimates dense multi-class 2D-3D correspondence maps between an input image and available 3D models. Pix2pose~\cite{2019Pix2pose} regresses pixel-wise 3D coordinates of objects from RGB images using 3D models without textures. EPOS~\cite{2020EPOS} represents objects by surface fragments which allows to handle symmetries, predicts a data-dependent number of precise 3D locations at each pixel, which establishes many-to-many 2D-3D correspondences, and utilizes an estimator for recovering poses of multiple object instances.

\subsubsection{3D point cloud-based methods}
\label{sec:3-1-2}

\begin{figure}[htbp]
\centering
\includegraphics[scale=0.35]{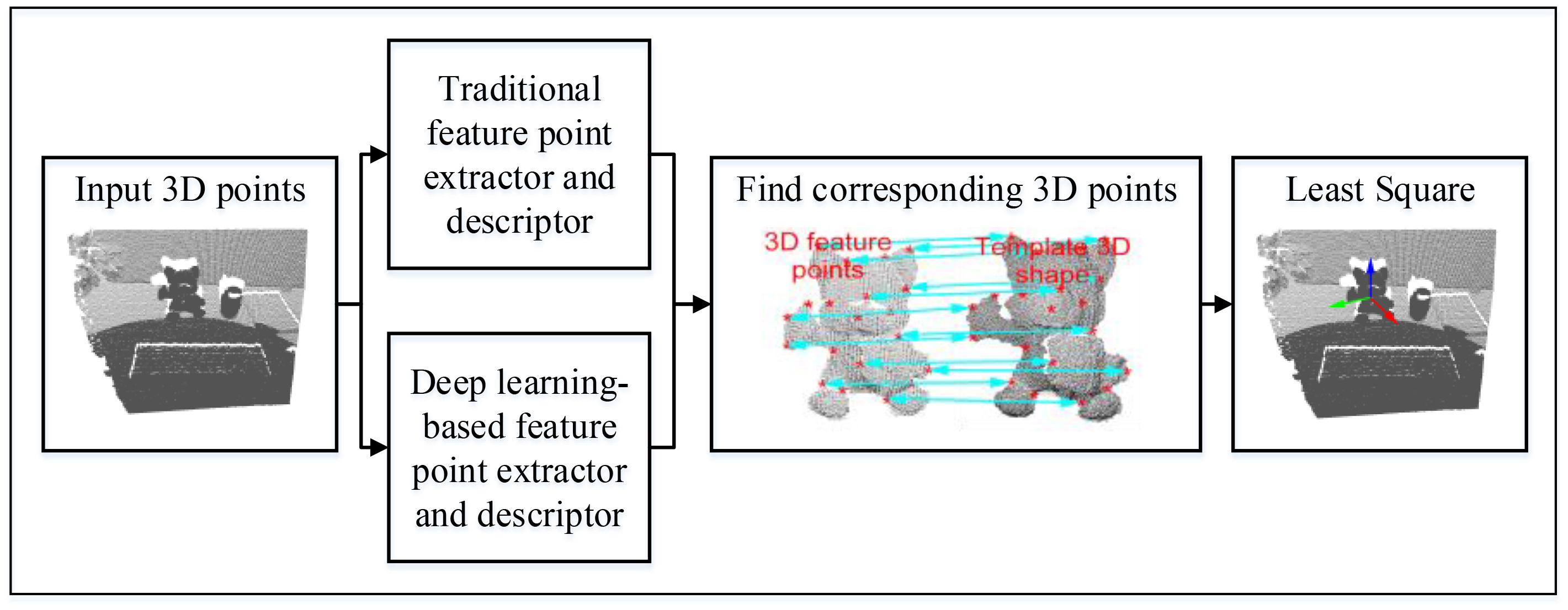}
\caption{Typical functional flow-chart of 3D correspondence-based 6D object pose estimation methods.}
\label{fig:3d-corres}
\end{figure}

Typical functional flow-chart of 3D correspondence-based 6D object pose estimation methods is illustrated in Fig.~\ref{fig:3d-corres}. When using the 3D point cloud lifted from the depth image, 3D geometric descriptors could be utilized for matching, which eliminates the influence of the texture. The 6D object pose could then be achieved by computing the transformations based on 3D-3D correspondences directly. The widely used 3D local shape descriptors, such as Spin Images~\cite{1997SpinImages}, 3D Shape Context~\cite{20043DSC}, FPFH~\cite{2009FPFH}, CVFH~\cite{2011CadRecognition6DOF}, SHOT~\cite{2014SHOT}, can be utilized to find correspondences between the object's partial 3D point cloud and full point cloud to obtain the 6D object pose. Some other 3D local descriptors could refer to the survey~\cite{2016Comprehensive3DLocalDesc}. However, this kind of methods require that the target objects have rich geometric features.

There also exist deep learning-based 3D descriptors~\cite{20173dmatch,20183dfeatnet} aiming at matching 3D points, which are representative and discriminative. 3DMatch~\cite{20173dmatch} is proposed to match 3D feature points using 3D voxel-based deep learning networks. 3DFeat-Net~\cite{20183dfeatnet} proposed a weakly supervised network that holistically learns a 3D feature detector and descriptor using only GPS/INS tagged 3D point clouds. Gojcic et al.~\cite{2019PerfectMatch} proposed 3DSmoothNet, which matches 3D point clouds with a siamese deep learning architecture and fully convolutional layers using a voxelized smoothed density value (SDV) representation. Yuan et al.~\cite{2020SelfSupervisedPSL} proposed a self-supervised learning method for descriptors in point clouds, which requires no manual annotation and achieves competitive performance. StickyPillars~\cite{2020Stickypillars} proposed an end-to-end trained 3D feature matching approach based on a graph neural network, and they perform context aggregation with the aid of transformer based multi-head self and cross attention.

\subsection{Template-based methods}
\label{sec:3-2}

Template-based 6D object pose estimation involves methods of finding the most similar template from the templates that are labeled with Ground Truth 6D object poses. In 2D case, the templates could be projected 2D images from known 3D models, and the objects within the templates have corresponding 6D object poses in the camera coordinate. The 6D object pose estimation problem is thus transformed into an image retrieval problem. In 3D case, the template could be the full point cloud of the target object. We need to find the best 6D pose that aligns the partial point cloud to the template and thus the 6D object pose estimation becomes a part-to-whole coarse registration problem. The methods of template-based methods are summarized in Table~\ref{tab:6d-template}.

\begin{table*}[htbp]
\caption{Summary of template-based 6D object pose estimation methods.}
    \begin{tabular}{ m{2.0cm} m{3.0cm} m{3.0cm} m{8.0cm} }
    \hline
    Methods & Descriptions & Traditional methods & Deep learning-based methods \\
    \hline
    2D image-based methods & Retrieve the template image that is most similar with the observed image & LineMod~\cite{2012ModelBased}, Hoda{\v{n}} et al.~\cite{2015DetectionAndFine} & AAE~\cite{2018Implicit}, PoseCNN~\cite{2018Posecnn}, SSD6D~\cite{2017SSD6D}, Deep-6DPose~\cite{2018Deep6DPose}, Liu et al.~\cite{2019Recovering6D}, CDPN~\cite{2019CDPN}, Tian et al.~\cite{2020Robust6DObject}, NOCS~\cite{2019NOCS}, LatentFusion~\cite{2020LatentFusion}, Chen et al.~\cite{2020LearningCASS} \\
    \hline
    3D point cloud-based methods & Find the pose that best aligns the observed partial 3D point cloud with the template full 3D model & Super4PCS~\cite{2014Super4PCS}, Go-ICP~\cite{2015Go-ICP} & PCRNet~\cite{2019PCRNet}, DCP~\cite{2019DeepCP}, PointNetLK~\cite{2019PointNetLK}, PRNet~\cite{2019PRNet},  DeepICP~\cite{2019DeepICP}, Sarode et al.~\cite{2019OneFrame}, TEASER~\cite{2020TEASER}, DGR~\cite{2020DeepGlobalReg}, G2L-Net~\cite{2020G2LNet}, Gao et al.~\cite{20206dObjPosReg}\\
    \hline
    \end{tabular}
\label{tab:6d-template}
\end{table*}

\subsubsection{2D image-based methods}
\label{sec:3-2-1}

\begin{figure}[htbp]
\centering
\includegraphics[scale=0.33]{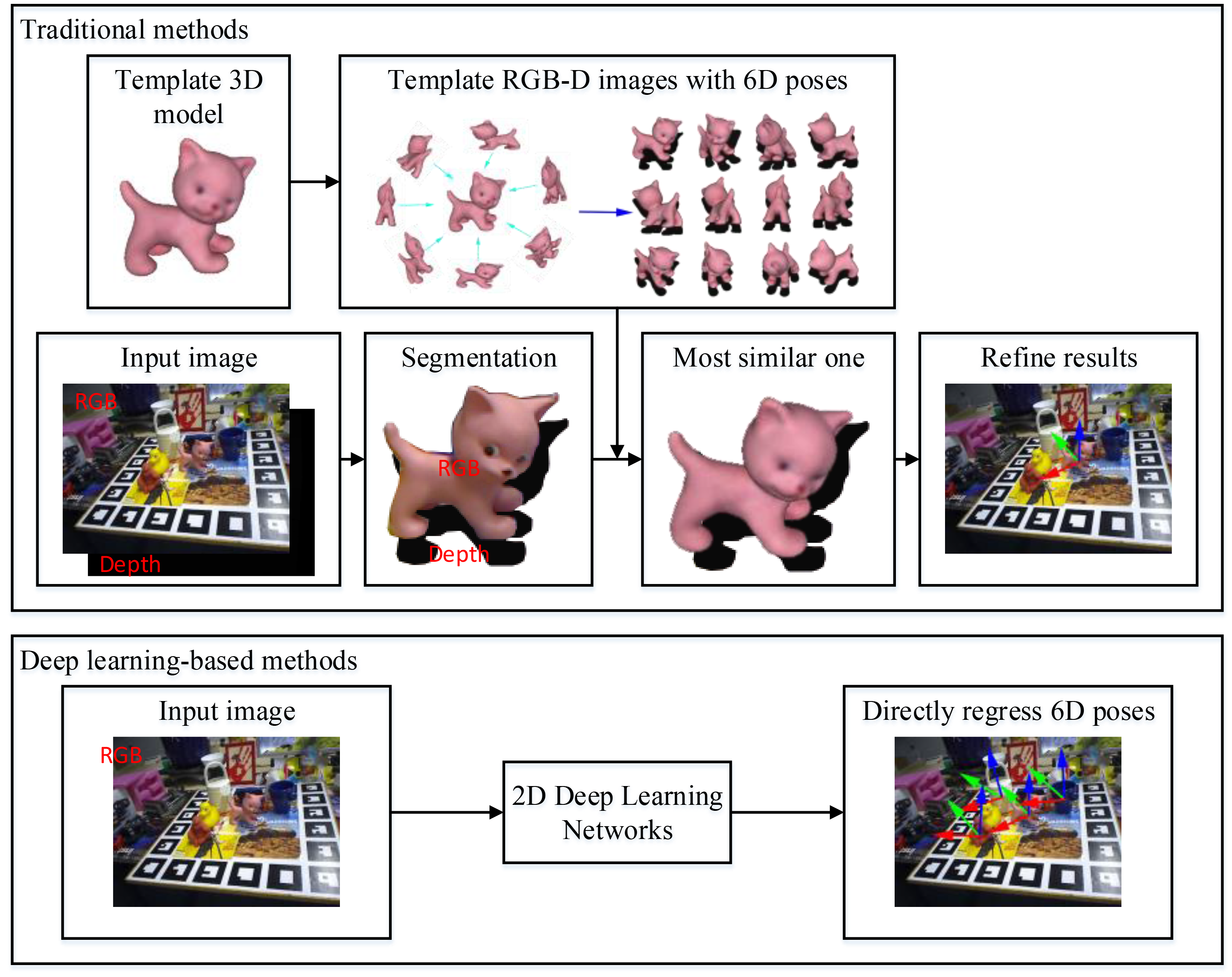}
\caption{Typical functional flow-chart of 2D template-based 6D object pose estimation methods. Data from the lineMod dataset~\protect\cite{2012ModelBased}.}
\label{fig:2d-template}
\end{figure}

Traditional 2D feature-based methods could be used to find the most similar template image and 2D correspondence-based methods could be utilized if discriminative feature points exist. Therefore, this kind of methods mainly aim at texture-less or non-texture objects that correspondence-based methods can hardly deal with. In these methods, the gradient information is usually utilized. Typical functional flow-chart of 2D template-based 6D object pose estimation methods is illustrated in Fig.~\ref{fig:2d-template}. Multiple images which are generated by projecting the existing complete 3D model from various angles are regarded as the templates. Hinterstoisser et al.~\cite{2012ModelBased} proposed a novel image representation by spreading image gradient orientations for template matching and represented a 3D object with a limited set of templates. The accuracy of the estimated pose was improved by taking into account the 3D surface normal orientations which are computed from the dense point cloud obtained from a dense depth sensor. Hoda{\v{n}} et al.~\cite{2015DetectionAndFine} proposed a method for the detection and accurate 3D localization of multiple texture-less and rigid objects depicted in RGB-D images. The candidate object instances are verified by matching feature points in different modalities and the approximate object pose associated with each detected template is used as the initial value for further optimization. There exist deep learning-based image retrieval methods~\cite{2016DeepImageRetrieval}, which could assist the template matching process. However, seldom of them were used in template-based methods and perhaps the number of templates is too small for deep learning methods to learn representative and discriminative features.

Above methods find the most similar template explicitly, and there also exist some implicitly ways. Sundermeyer et al.~\cite{2018Implicit} proposed Augmented Autoencoders (AAE), which learns the 3D orientation implicitly. Thousands of template images are rendered from a full 3D model and these template images are encoded into a codebook. The input image will be encoded into a new code and matched with the codebook to find the most similar template image, and the 6D object pose is thus obtained.

There also exist methods~\cite{2018Posecnn,2018Deep6DPose,2019Recovering6D} that directly estimate the 6D pose of the target object from the input image, which can be regarded as finding the most similar image from the pre-trained and labeled images implicitly. Different from correspondence-based methods, this kind of method learns the immediate mapping from an input image to a parametric representation of the pose, and the 6D object pose can thus be estimated combined with object detection~\cite{2018SurveyJoint}. Yu et al.~\cite{2018Posecnn} proposed PoseCNN for direct 6D object pose estimation. The 3D translation of an object is estimated by localizing the center in the image and predicting the distance from the camera, and the 3D rotation is computed by regressing a quaternion representation. Kehl et al.~\cite{2017SSD6D} presented a similar method by making use of the SSD network. Do et al.~\cite{2018Deep6DPose} proposed an end-to-end deep learning framework named Deep-6DPose, which jointly detects, segments, and recovers 6D poses of object instances form a single RGB image. They extended the instance segmentation network Mask R-CNN~\cite{2017MaskRCNN} with a pose estimation branch to directly regress 6D object poses without any post-refinements. Liu et al.~\cite{2019Recovering6D} proposed a two-stage CNN architecture which directly outputs the 6D pose without requiring multiple stages or additional post-processing like PnP. They transformed the pose estimation problem into a classification and regression task. CDPN~\cite{2019CDPN} proposed the Coordinates-based Disentangled Pose Network (CDPN), which disentangles the pose to predict rotation and translation separately. Tian et al.~\cite{2020Robust6DObject} also proposed a discrete-continuous formulation for rotation regression to resolve this local-optimum problem. They uniformly sample rotation anchors in $SO(3)$, and predict a constrained deviation from each anchor to the target.

There also exist methods that build a latent representation for category-level objects. This kind of methods can also be treated as the template-based methods, and the template could be implicitly built from multiple images. NOCS~\cite{2019NOCS}, LatentFusion~\cite{2020LatentFusion} and Chen et al.~\cite{2020LearningCASS} are the representative methods.

\subsubsection{3D point cloud-based methods}
\label{sec:3-2-2}

\begin{figure}[htbp]
\centering
\includegraphics[scale=0.3]{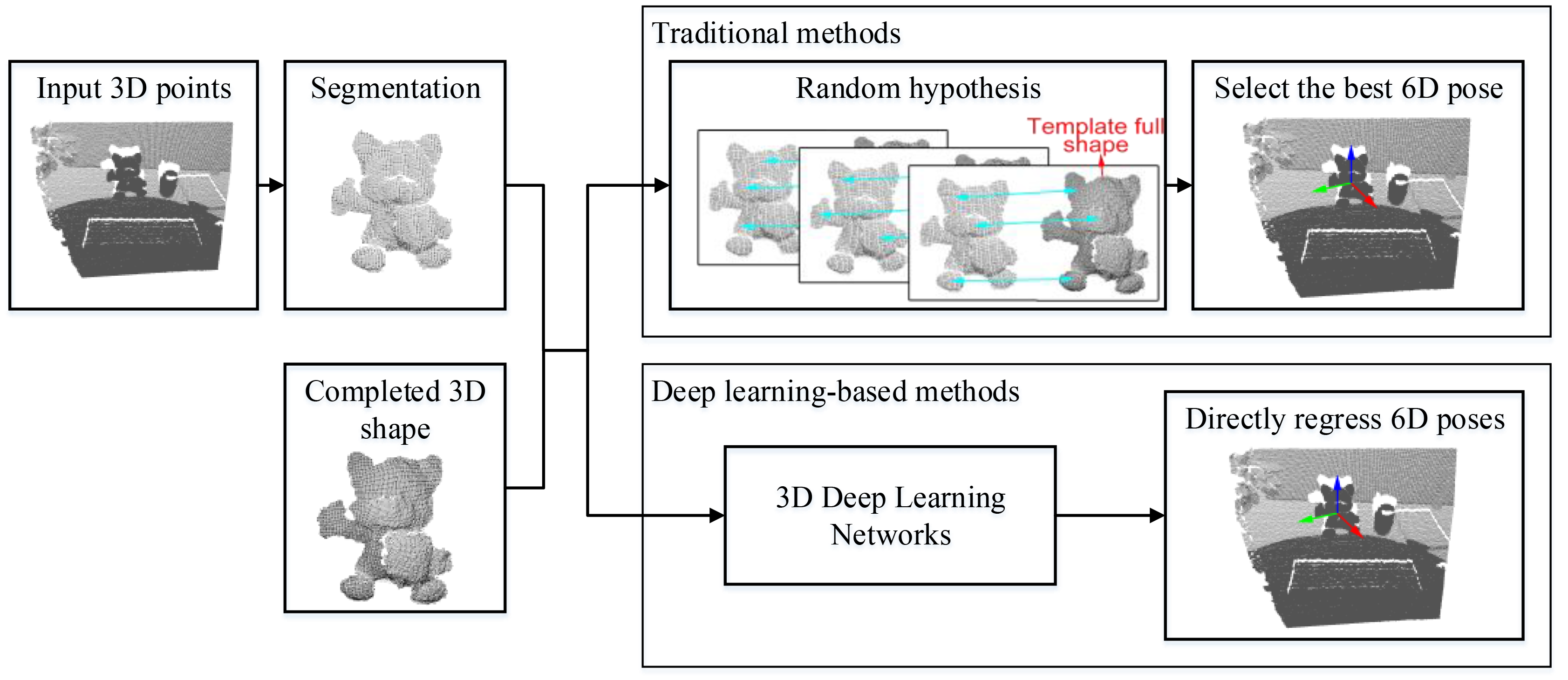}
\caption{Typical functional flow-chart of 3D template-based 6D object pose estimation methods.}
\label{fig:3d-template}
\end{figure}

Typical functional flow-chart of 3D template-based 6D object pose estimation methods is illustrated in Fig.~\ref{fig:3d-template}. Traditional partial registration methods aim at finding the 6D transformation that best aligns the partial point cloud to the full point cloud. Various global registration methods~\cite{2014Super4PCS,2015Go-ICP,2016FGR} exist which afford large variations of initial poses and are robust with large noise. However, this kind of method is time-consuming. Most of these methods utilize local registration methods such as the iterative closest points(ICP) algorithm~\cite{1992ICP} to refine the results. This part can refer to some review papers~\cite{2013Registration3DSurvey,2014Survey3DRegistration}.

Some deep learning-based methods also exist, which can accomplish the partial registration task in an efficient way. These methods consume a pair of point clouds, extract representative and discriminative features from 3D deep learning networks, and regress the relative 6D transformations between the pair of point clouds. PCRNet~\cite{2019PCRNet}, DCP~\cite{2019DeepCP}, PointNetLK~\cite{2019PointNetLK}, PRNet~\cite{2019PRNet}, DeepICP~\cite{2019DeepICP}, Sarode et al.~\cite{2019OneFrame}, TEASER~\cite{2020TEASER} and DGR~\cite{2020DeepGlobalReg} are the representative methods and readers could refer to the recent survey~\cite{2020DeepAlignmentSurvey}. There also exist methods~\cite{2020G2LNet,20206dObjPosReg} that directly regress the 6D object pose from the partial point cloud. G2L-Net~\cite{2020G2LNet} extracts the coarse object point cloud from the RGB-D image by 2D detection, and then conducts translation localization and rotation localization. Gao et al.~\cite{20206dObjPosReg} conducts 6D object pose regression via supervised learning on point clouds.

\subsection{Voting-based methods}
\label{sec:3-3}

Voting-based methods mean that each pixel or 3D point contributes to the 6D object pose estimation by providing one or more votes. We roughly divide voting methods into two kinds, which are indirectly voting methods and directly voting methods. Indirectly voting methods mean that each pixel or 3D point vote for some feature points, which affords 2D-3D correspondences or 3D-3D correspondences. Directly voting methods mean that each pixel or 3D point vote for a certain 6D object coordinate or pose. These methods are summarized in Table~\ref{tab:6d-voting}.

\begin{table*}[htbp]
\caption{Summary of voting-based 6D object pose estimation methods.}
    \begin{tabular}{ m{2.0cm} m{3.0cm} m{5.5cm} m{5.5cm} }
    \hline
    Methods & Descriptions & 2D image-based methods & 3D point cloud-based methods \\
    \hline
    Indirect voting methods & Voting for correspondence-based methods & PVNet~\cite{2019PVNet}, Yu et al.~\cite{20206DoFViaDPVL} & PVN3D~\cite{2020PVN3D}, YOLOff~\cite{2020YOLOff}, 6-PACK~\cite{20196-PACK}\\
    \hline
    Direct voting methods & Voting for template-based methods & Brachmann et al.~\cite{2014Learning6DObject}, Tejani et al.~\cite{2014LatentClassHough}, Crivellaro et al.~\cite{2017Robust3D}, PPF~\cite{2012PPF} & DenseFusion~\cite{2019DenseFusion}, MoreFusion~\cite{2020MoreFusion} \\
    \hline
    \end{tabular}
\label{tab:6d-voting}
\end{table*}

\subsubsection{Indirect voting methods}
\label{sec:3-3-1}

\begin{figure}[htbp]
\centering
\includegraphics[scale=0.3]{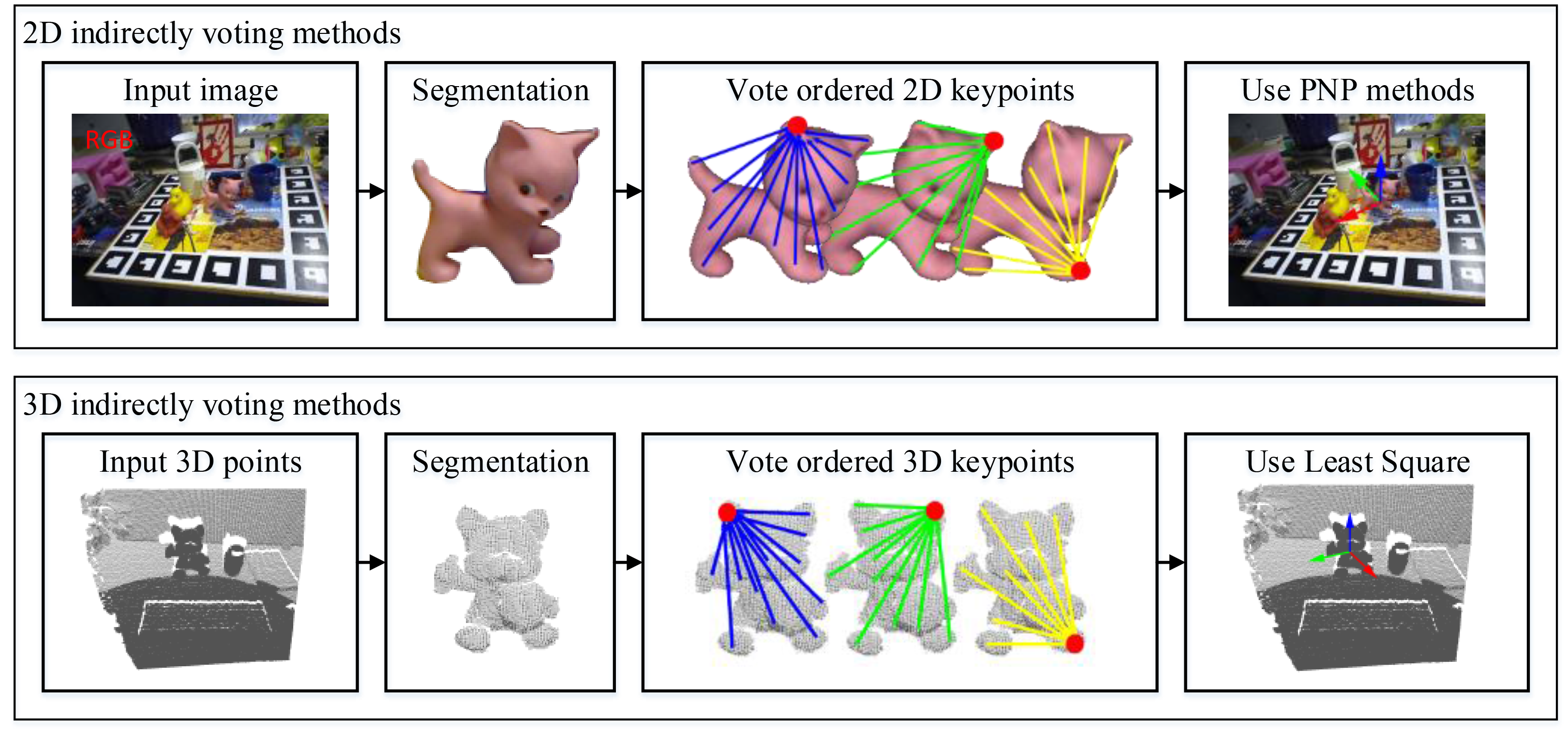}
\caption{Typical functional flow-chart of indirect voting-based object pose estimation methods.}
\label{fig:vote4points}
\end{figure}

This kind of methods can be regarded as voting for correspondence-based methods. In 2D case, 2D feature points are voted and 2D-3D correspondences could be achieved. In 3D case, 3D feature points are voted and 3D-3D correspondences between the observed partial point cloud and the canonical full point cloud could be achieved. Most of this kind of methods utilize deep learning methods for their strong feature representation capabilities in order to predict better votes. Typical functional flow-chart of indirect voting-based 6D object pose estimation methods is illustrated in Fig.~\ref{fig:vote4points}.

In 2D case, PVNet~\cite{2019PVNet} votes projected 2D feature points and then finds the corresponding 2D-3D correspondences to compute the 6D object pose. Yu et al.~\cite{20206DoFViaDPVL} proposed a method which votes 2D positions of the object keypoints from vector-fields. They develop a differentiable proxy voting loss (DPVL) which mimics the hypothesis selection in the voting procedure. In 3D case, PVN3D~\cite{2020PVN3D} votes 3D keypoints, and can be regarded as a variation of PVNet~\cite{2019PVNet} in 3D domain. YOLOff~\cite{2020YOLOff} utilizes a classification CNN to estimate the object's 2D location in the image from local patches, followed by a regression CNN trained to predict the 3D location of a set of keypoints in the camera coordinate system. The 6D object pose is then achieved by minimizing a registration error. 6-PACK~\cite{20196-PACK} predicts a handful of ordered 3D keypoints for an object based on the observation that inter-frame motion of an object instance can be estimated through keypoint matching. This method achieves category-level 6D object pose tracking on RGB-D data.

\subsubsection{Direct voting methods}
\label{sec:3-3-2}

\begin{figure}[htbp]
\centering
\includegraphics[scale=0.33]{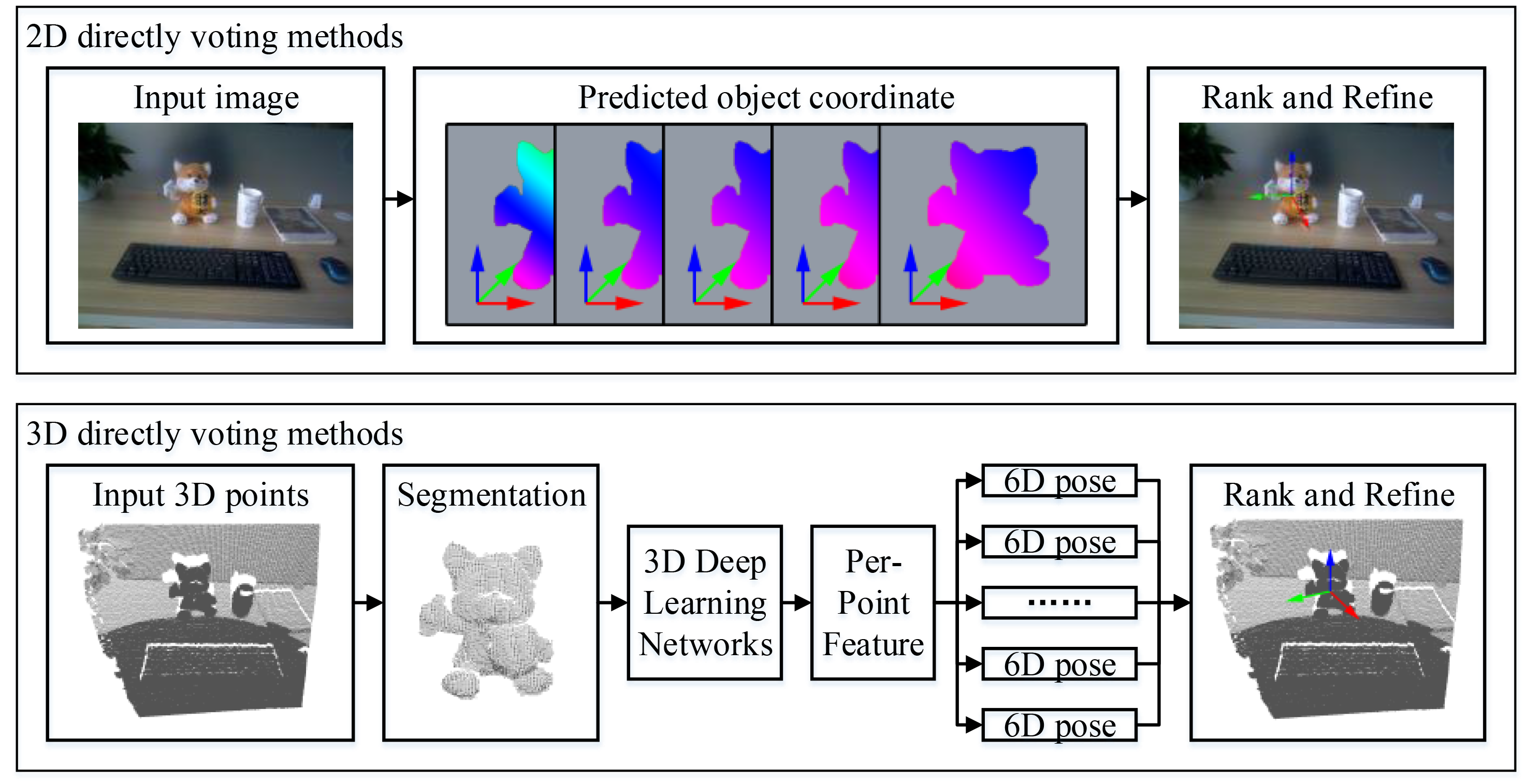}
\caption{Typical functional flow-chart of direct voting-based 6D object pose estimation methods.}
\label{fig:vote4pose}
\end{figure}

This kind of methods can be regarded as voting for template-based methods if we treat the voted object pose or object coordinate as the most similar template. Typical functional flow-chart of direct voting-based 6D object pose estimation methods is illustrated in Fig.~\ref{fig:vote4pose}.

In 2D case, this kind of methods are mainly used for computing the poses of objects with occlusions. For these objects, the local evidence in the image restricts the possible outcome of the desired output, and every image patch is thus usually used to cast a vote about the 6D object pose. Brachmann et al.~\cite{2014Learning6DObject} proposed a learned, intermediate representation in the form of a dense 3D object coordinate labelling paired with a dense class labelling. Each object coordinate prediction defines a 3D-3D correspondence between the image and the 3D object model, and the pose hypotheses are generated and refined to obtain the final hypothesis. Tejani et al.~\cite{2014LatentClassHough} trained a Hough forest for 6D pose estimation from an RGB-D image. Each tree in the forest maps an image patch to a leaf which stores a set of 6D pose votes.

In 3D case, Drost et al.~\cite{2010DrostModel} proposed the Point Pair Features (PPF) to recover the 6D pose of objects from a depth image. A point pair feature contains information about the distance and normals of two arbitrary 3D points. PPF has been one of the most successful 6D pose estimation method as an efficient and integrated alternative to the traditional local and global pipelines. Hodan et al.~\cite{2018BenchmarkSurvey} proposed a benchmark for 6D pose estimation of a rigid object from a single RGB-D input image, and a variation of PPF~\cite{2018Vidal6DPose} won the 2018 SIXD challenge.

Deep learning-based methods also assist the directly voting methods. DenseFusion~\cite{2019DenseFusion} utilizes a heterogeneous architecture that processes the RGB and depth data independently and extracts pixel-wise dense feature embeddings. Each feature embedding votes a 6D object pose and the best prediction is adopted. They further proposed an iterative pose refinement procedure to refine the predicted 6D object pose. MoreFusion~\cite{2020MoreFusion} conducts an object-level volumetric fusion and performs point-wise volumetric pose prediction that exploits volumetric reconstruction and CNN feature extraction from the image observation. The object poses are then jointly refined based on geometric consistency among objects and impenetrable space.

\subsection{Comparisons and discussions}
\label{sec:3-4}

In this section, we mainly review the methods based on the RGB-D image, since 3D point cloud-based 6D object pose estimation could be regarded as a registration or alignment problem where some surveys~\cite{2013Registration3DSurvey,2014Survey3DRegistration} exist. The related datasets, evaluation metrics and comparisons are presented.

\subsubsection{Datasets and evaluation metrics}
\label{sec:3-4-1}

There exist various benchmarks~\cite{20186DBenchmark} for 6D pose estimation, such as LineMod~\cite{2012ModelBased}, IC-MI/IC-BIN dataset~\cite{2014LatentClassHough}, T-LESS dataset~\cite{2017Tless}, RU-APC dataset~\cite{2015RutgersAPC}, and YCB-Video~\cite{2018Posecnn}, etc. Here we only reviewed the most widely used LineMod~\cite{2012ModelBased} dataset and YCB-Video~\cite{2018Posecnn} dataset. LineMod~\cite{2012ModelBased} provides manual annotations for around 1,000 images for each of the 15 objects in the dataset. Occlusion Linemod~\cite{2014Learning6DObject} contains more examples where the objects are under occlusion. YCB-Video~\cite{2018Posecnn} contains a subset of 21 objects and comprises 133,827 images. These datasets are widely evaluated aiming at various kinds of methods.

The 6D object pose can be represented by a $4 \times 4$ matrix ${\mathop{\rm P}\nolimits}  = [R,t;0,1]$, where $R$ is a $3 \times 3$ rotation matrix and $t$ is a $3 \times 1$ translation vector. The rotation matrix could also be represented as quaternions or angle-axis representation. Direct comparison of the variances between the values can not provide intuitive visual understandings. The commonly used metrics are the Average Distance of Model Points (ADD)~\cite{2012ModelBased} for non-symmetric objects and the average closest point distances (ADD-S)~\cite{2018Posecnn} for symmetric objects.

Given a 3D model $M$, the ground truth rotation $R$ and translation $T$, and the estimated rotation $\hat R$ and translation $\hat T$, ADD means the average distance of all model points $x$ from their transformed versions. The 6D object pose is considered to be correct if the average distance is smaller than a predefined threshold.
\begin{equation}
    {e_{ADD}} = \mathop {avg}\limits_{x \in M} \left\| {(Rx + T) - (\hat Rx + \hat T)} \right\|.
\end{equation}
ADD-S~\cite{2018Posecnn} is an ambiguity-invariant pose error metric which takes both symmetric and non-symmetric objects into an overall evaluation. Given the estimated pose ${\rm{[\hat R|\hat T]}}$ and the ground truth pose ${\rm{[R|T]}}$, ADD-S calculates the mean distance from each 3D model point transformed by ${\rm{[\hat R|\hat T]}}$ to its closest point on the target model transformed by ${\rm{[R|T]}}$.

Aim at the LineMOD dataset, ADD is used for asymmetric objects and ADD-S is used for symmetric objects. The threshold is usually set as 10$\%$ of the model diameter. Aiming at the YCB-Video dataset, the commonly used evaluation metric is the ADD-S metric. The percentage of ADD-S smaller than 2cm ($<$2cm) is often used, which measures the predictions under the minimum tolerance for robotic manipulation. In addition, the area under the ADD-S curve (AUC) following PoseCNN~\cite{2018Posecnn} is also reported and the maximum threshold of AUC is set to be 10cm.

\begin{table*}[htbp]
\centering
\caption{Accuracies of AUC and ADD-S metrics on YCB-video dataset.}
 \begin{tabular}{m{2.5cm} m{7.5cm} m{2.0cm} m{2.0cm}}
    \toprule
    Category & Method & AUC & ADD-S ($<$2cm) \\
    \midrule
    Corre-based & Heatmaps~\cite{2018MakingDeepHeatmaps} & 72.8 & 53.1\\
    \hline
    \multirow{5}{2.5cm}{Template-based} & PoseCNN~\cite{2018Posecnn}+ICP & 61.0 & 73.8 \\
    & PoseCNN~\cite{2018Posecnn}+ICP & 93.0 & 93.2 \\
    & Castro et al.~\cite{2019Accurate6DByRecon} & 67.52 & 47.09 \\
    & PointFusion~\cite{2018PointFusion} & 83.9 & 74.1 \\
    & MaskedFusion~\cite{2019MaskedFusion} & 93.3 & 97.1 \\
    \hline
    \multirow{2}{2.5cm}{Voting-based} & DenseFusion~\cite{2019DenseFusion}(per-pixel) & 91.2 & 95.3 \\
     & DenseFusion~\cite{2019DenseFusion}(iterative) & 93.1 & 96.8\\
    \bottomrule
    \end{tabular}
\label{tab:posecompYCB}
\end{table*}

\begin{table*}[htbp]
\centering
\caption{Accuracies of methods using ADD(-S) metric on LineMOD and Occlusion LineMOD dataset. Refine means methods such as ICP or DeepIM. IR is short for iterative refinement.}
 \begin{tabular}{ m{2.5cm} m{7.5cm} m{2.0cm} m{2.0cm}}
    \toprule
    Category &  Method & LineMOD & Occlusion \\
    \midrule
    \multirow{10}{2.5cm}{Correspondence-based methods}
    & BB8~\cite{2017BB8} & 43.6 & - \\
    & BB8~\cite{2017BB8}+Refine & 62.7 & - \\
    & Tekin et al.~\cite{2018RealTime} & 55.95 & 6.42 \\
    & Heatmaps~\cite{2018MakingDeepHeatmaps} & - & 25.8 \\
    & Heatmaps~\cite{2018MakingDeepHeatmaps}+Refine & - & 30.4 \\
    & Hu et al.~\cite{2018SegmentationDriven} & - & 26.1 \\
    & Pix2pose~\cite{2019Pix2pose} & 72.4 & 32.0 \\
    & DPOD~\cite{2019DPOD} & 82.98 & 32.79 \\
    & DPOD~\cite{2019DPOD}+Refine & 95.15 & 47.25 \\
    & HybridPose~\cite{2020HybridPose} & 94.5 & 79.2 \\
    \hline
    \multirow{10}{2.5cm}{Template-based methods}
    & SSD-6D~\cite{2017SSD6D} & 2.42 & - \\
    & SSD-6D~\cite{2017SSD6D}+Refine & 76.7 & 27.5 \\
    & AAE~\cite{2018Implicit} & 31.41 & - \\
    & AAE~\cite{2018Implicit}+Refine & 64.7 & - \\
    & Castro et al.~\cite{2019Accurate6DByRecon} & 59.32 & - \\
    & PoseCNN~\cite{2018Posecnn} & 62.7 & 6.42 \\
    & PoseCNN~\cite{2018Posecnn}+Refine & 88.6 & 78.0 \\
    & CDPN~\cite{2019CDPN} & 89.86 & - \\
    & Tian et al.~\cite{2020Robust6DObject} & 92.87 & - \\
    & MaskedFusion~\cite{2019MaskedFusion} & 97.3 & - \\
    \hline
    \multirow{13}{2.5cm}{Voting-based methods}
    & Brachmann et al.~\cite{2016UncertaintyDriven6D} & 32.3 & - \\
    & Brachmann et al.~\cite{2016UncertaintyDriven6D}+Refine & 50.2 & - \\
    & PVNet~\cite{2019PVNet} & 86.27 & 40.8 \\
    & DenseFusion~\cite{2019DenseFusion}(per-pixel) & 86.2 \\
    & DenseFusion~\cite{2019DenseFusion}(iterative) & 94.3 \\
    & DPVL~\cite{20206DoFViaDPVL} & 91.5 & 43.52 \\
    & YOLOff~\cite{2020YOLOff} & 94.2 & - \\
    & YOLOff~\cite{2020YOLOff}+Refine & 98.1 & - \\
    & PVN3D~\cite{2020PVN3D} & 95.1 & - \\
    & P$^{2}$GNet~\cite{2019P2GNet} & 96.2 & - \\
    & P$^{2}$GNet~\cite{2019P2GNet}+Refine & 97.4 & - \\
    & PointPoseNet~\cite{2019PointPoseNet} & 96.3 & 52.6 \\
    & PointPoseNet~\cite{2019PointPoseNet}+Refine & - & 75.1 \\
    \bottomrule
    \end{tabular}
\label{tab:posecompLineMOD}
\end{table*}

\subsubsection{Comparisons and discussions}
\label{sec:3-4-2}

6D object pose estimation plays a pivotal role in robotic and augment reality areas. Various methods exist with different inputs, precision, speed, advantages and disadvantages. Aiming at robotic grasping tasks, the practical environment, the available input data, the available hardware setup, the target objects to be grasped, the task requirements should be analyzed first to decide which kinds of methods to use. The above mentioned three kinds of methods deal with different kinds of objects. Generally, when the target object has rich texture or geometric details, the correspondence-based method is a good choice. When the target object has weak texture or geometric detail, the template-based method is a good choice. When the object is occluded and only partial surface is visible, or the addressed object ranges from specific objects to category-level objects, the voting-based method is a good choice. Besides, the three kinds of methods all have 2D inputs, 3D inputs or mixed inputs. The results of methods with RGB-D images as inputs are summarized in Table~\ref{tab:posecompYCB} on the YCB-Video dataset, and Table~\ref{tab:posecompLineMOD} on the LineMOD and Occlusion LineMOD datasets. All recent methods on LineMOD achieve high accuracy since there's little occlusion. When there exist occlusions, correspondence-based and voting-based methods perform better than template-based methods. The template-based methods are more like a direct regression problem, which highly depend on the global feature extracted. Whereas, correspondence-based and voting-based methods utilize the local parts information and constitute local feature representations.

There exist some challenges for nowadays 6D object pose estimation methods. The first challenge lies in that current methods show obvious limitations in cluttered scenes in which occlusions usually occur. Although the state-of-the-art methods achieve high accuracies on the Occlusion LineMOD dataset, they still could not afford severe occluded cases since this situation may cause ambiguities even for human beings. The second one is the lack of sufficient training data, as the sizes of the datasets presented above are relatively small. Nowadays deep learning methods show poor performance on objects which do not exist in the training datasets and perhaps the simulated datasets could be one solution. Although some category-level 6D object pose methods~\cite{2019NOCS,2020LatentFusion,2020LearningCASS} emerged recently, they still can not handle large number of categories.

\section{Grasp Estimation}
\label{sec:4}

Grasp estimation means estimating the 6D gripper pose in the camera coordinate. As mentioned before, the grasp can be categorized into 2D planar grasp and 6DoF grasp. For 2D planar grasp, where the grasp is constrained from one direction, the 6D gripper pose could be simplified into a 3D representation, which includes the 2D in-plane position and 1D rotation angle, since the height and the rotations along other axes are fixed. For 6DoF grasp, the gripper can grasp the object from various angles and the 6D gripper pose is essential to conduct the grasp. In this section, methods of 2D planar grasp and 6DoF grasp are presented in detail.

\subsection{2D planar grasp}
\label{4-1}

\begin{figure}[htbp]
\centering
\includegraphics[scale=0.4]{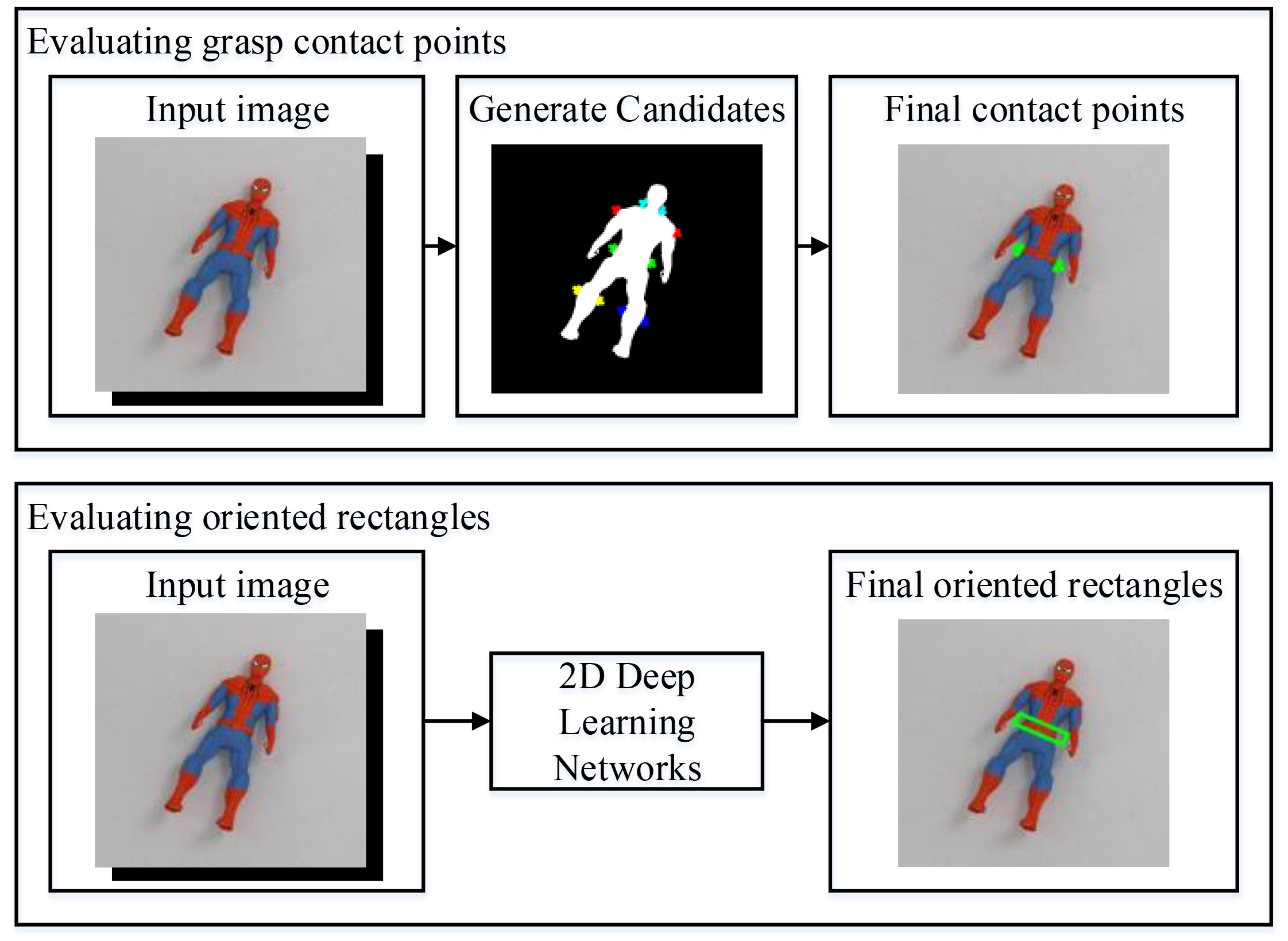}
\caption{Typical functional flow-chart of 2D planar grasp methods. Data from the JACQUARD dataset~\protect\cite{2018JacquardDataset}.}
\label{fig:2dplanargrasp}
\end{figure}

Methods of 2D planar grasp can be divided into methods of evaluating grasp contact points and methods of evaluating oriented rectangles. In 2D planar grasp, the grasp contact points can uniquely define the gripper's grasp pose, which is not the situation in 6DoF grasp. The 2D oriented rectangles can also uniquely define the gripper's grasp pose. These methods are summarized in Table~\ref{tab:2d-grasp} and typical functional flow-chart is illustrated in Fig.~\ref{fig:2dplanargrasp}.

\begin{table*}[htbp]
\caption{Summary of 2D planar grasp estimation methods.}
    \begin{tabular}{ m{3.0cm} m{3.0cm} m{10.0cm} }
    \hline
    Methods & Traditional methods & Deep learning-based methods \\
    \hline
    Methods of evaluating grasp contact points & Domae et al.~\cite{2014FastGraspability} & Zeng et al.~\cite{2018RoboticPick}, Mahler et al.~\cite{2017DexNet2}, Cai et al.~\cite{2019MetaGrasp}, GG-CNN~\cite{2018ClosingTheLoop}, MVP~\cite{2019MultiViewPicking}, Wang et al.~\cite{2019EfficientFCNN} \\
    \hline
    Methods of evaluating oriented rectangles & Jiang et al.~\cite{2011EfficientGrasping}, Vohra et al.~\cite{2019RealTimeGraspPose} & Lenz et al.~\cite{2015DeepLearnDetect}, Pinto and Gupta~\cite{2016SupersizingSelfSuper}, Park and Chun~\cite{2018ClassificationBasedGrasp}, Redmon and Angelova~\cite{2015RealGrasp}, Zhang et al.~\cite{2017RobustRobotGrasp}, Kanan~\cite{2017RoboticGrasp}, Kumra et al.~\cite{2019AntipodalRoboticGrasping}, Zhang et al.~\cite{2018ROIBased}, Guo et al.~\cite{2017AHybridDeep}, Chu et al.~\cite{2018RealWorldMultiObj}, Park et al.~\cite{2018RealTimeHighly}, Zhou et al.~\cite{2018FullyConvolutionalGrasp}, Depierre et al.~\cite{2020OptimizingCorrelated} \\
    \hline
    \end{tabular}
\label{tab:2d-grasp}
\end{table*}

\subsubsection{Methods of evaluating grasp contact points}
\label{4-1-1}

This kind of methods first sample candidate grasp contact points, and use analytical methods or deep learning-based methods to evaluate the possibility of a successful grasp, which are classification-based methods. Empirical methods of robotic grasping are performed based on the premise that certain prior knowledge, such as object geometry, physics models, or force analytic, are known. The grasp database usually covers a limited amount of objects, and empirical methods will face difficulties in dealing with unknown objects. Domae et al.~\cite{2014FastGraspability} presented a method that estimates graspability measures on a single depth map for grasping objects randomly placed in a bin. Candidate grasp regions are first extracted and the graspability is computed by convolving one contact region mask image and one collision region mask image. Deep learning-based methods could assists in evaluating the grasp qualities of candidate grasp contact points. Mahler et al.~\cite{2017DexNet2} proposed DexNet 2.0, which plans robust grasps with synthetic point clouds and analytic grasping metrics. They first segment the current points of interests from the depth image, and multiple candidate grasps are generated. The grasp qualities are then measured using the Grasp Quality-CNN network, and the one with the highest quality will be selected as the final grasp. Their database have more than 50k grasps, and the grasp quality measurement network achieved relatively satisfactory performance.

Deep learning-based methods could also assist in estimating the most probable grasp contact points through estimating pixel-wise grasp affordances. Robotic affordances~\cite{2018AffordanceNet,2019LearningGraspAffordance,2019DetectingRoboticAffordances} usually aim to predict affordances of the object parts for robot manipulation, which are more like a segmentation problem. However, there exist some methods~\cite{2018RoboticPick,2019MetaGrasp} that predict pixel-wise affordances with respect to the grasping primitive actions. These methods generate grasp qualities for each pixel, and the pair of points with the highest affordance value is executed. Zeng et al.~\cite{2018RoboticPick} proposed a method which infers dense pixel-wise probability maps of the affordances for four different grasping primitive actions through utilizing fully convolutional networks. Cai et al.~\cite{2019MetaGrasp} presented a pixel-level affordance interpreter network, which learns antipodal grasp patterns based on a fully convolutional residual network similar with Zeng et al.~\cite{2018RoboticPick}. Both of these two methods do not segment the target object and predict pixel-wise affordance maps for each pixels. This is a way which directly estimate grasp qualities without sampling grasp candidates. Morrison et al.~\cite{2018ClosingTheLoop} proposed the Generative Grasping Convolutional Neural Network (GG-CNN), which predicts the quality and pose of grasps at every pixel. Further, Morrison et al.~\cite{2019MultiViewPicking} proposed a Multi-View Picking (MVP) controller, which uses an active perception approach to choose informative viewpoints based on a distribution of grasp pose estimates. They utilized the real-time GG-CNN~\cite{2018ClosingTheLoop} for visual grasp detection. Wang et al.~\cite{2019EfficientFCNN} proposed a fully convolution neural network which encodes the origin input images to features and decodes these features to generate robotic grasp properties for each pixel. Unlike classification-based methods for generating multiple grasp candidates through neural network, their pixel-wise implementation directly predicts multiple grasp candidates through one forward propagation.

\subsubsection{Methods of evaluating oriented rectangles}
\label{4-1-2}

Jiang et al.~\cite{2011EfficientGrasping} first proposed to use an oriented rectangle to represent the gripper configuration and they utilized a two-step procedure, which first prunes the search space using certain features that are fast to compute and then uses advanced features to accurately select a good grasp. Vohra et al.~\cite{2019RealTimeGraspPose} proposed a grasp estimation strategy which estimates the object contour in the point cloud and predicts the grasp pose along with the object skeleton in the image plane. Grasp rectangles at each skeleton point are estimated, and point cloud data corresponding to the grasp rectangle part and the centroid of the object is used to decide the final grasp rectangle. Their method is simple and needs no grasp configuration sampling steps.

Aiming at the oriented rectangle-based grasp configuration, deep learning methods are gradually applied in three different ways, which are classification-based methods, regression-based methods and detection-based methods. Most of these methods utilize a five dimensional representation~\cite{2015DeepLearnDetect} for robotic grasps, which are rectangles with a position, orientation and size: (x,y,$\theta$,h,w).

Classification-based methods train classifiers to evaluate candidate grasps, and the one with the highest score will be selected. Lenz et al.~\cite{2015DeepLearnDetect} is the first to apply deep learning methods to robotic grasping. They presented a two-step cascaded system with two deep networks, where the top detection results from the first are re-evaluated by the second. The first network produces a small set of oriented rectangles as candidate grasps, which will be axis aligned. The second network ranks these candidates using features extracted from the color image, the depth image and surface normals. The top-ranked rectangle is selected and the corresponding grasp is executed. Pinto and Gupta~\cite{2016SupersizingSelfSuper} predicted grasping locations by sampling image patches and predicting the grasping angle. They trained a CNN-based classifier to estimate the grasp likelihood for different grasp directions given an input image patch. Park and Chun~\cite{2018ClassificationBasedGrasp} proposed a classification based robotic grasp detection method with multiple-stage spatial transformer networks (STN). Their method allows partial observation for intermediate results such as grasp location and orientation for a number of grasp configuration candidates. The procedure of classification-based methods is straightforward, and the accuracy is relatively high. However, these methods tend to be quite slow.

Regression-based methods train a model to yield grasp parameters for location and orientation directly, since a uniform network would perform better than the two-cascaded system~\cite{2015DeepLearnDetect}. Redmon and Angelova~\cite{2015RealGrasp} proposed a larger neural network, which performs a single-stage regression to obtain graspable bounding boxes without using standard sliding window or region proposal techniques. Zhang et al.~\cite{2017RobustRobotGrasp} utilized a multi-modal fusion architecture which combines RGB features and depth features to improve the grasp detection accuracy. Kumra and Kanan~\cite{2017RoboticGrasp} utilized deep neural networks like ResNet~\cite{2016ResNet} and further increased the performances in grasp detection. Kumra et al.~\cite{2019AntipodalRoboticGrasping} proposed a novel Generative Residual Convolutional Neural Network (GR-ConvNet) model that can generate robust antipodal grasps from a n-channel image input. Rather than regressing the grasp parameters globally, some methods utilized a ROI (Region of Interest)-based or pixel-wise way. Zhang et al.~\cite{2018ROIBased} utilized ROIs in the input image and regressed the grasp parameters based on ROI features.

Detection-based methods utilize the reference anchor box, which are used in some deep learning-based object detection algorithms~\cite{2015FasterRCNN,2016SSD,2016YOLO}, to assist the generation and evaluation of candidate grasps. With the prior knowledge on the size of the expected grasps, the regression problem is simplified~\cite{2020OptimizingCorrelated}. Guo et al.~\cite{2017AHybridDeep} presented a hybrid deep architecture combining the visual and tactile sensing. They introduced the reference box which is axis aligned. Their network produces a quality score and an orientation as classification between discrete angle values. Chu et al.~\cite{2018RealWorldMultiObj} proposed an architecture that predicts multiple candidate grasps instead of a single outcome and transforms the orientation regression to a classification task. The orientation classification contains the quality score and therefore their network predicts both grasp regression values and discrete orientation classification score. Park et al.~\cite{2018RealTimeHighly} proposed a rotation ensemble module (REM) for robotic grasp detection using convolutions that rotates network weights. Zhou et al.~\cite{2018FullyConvolutionalGrasp} designed an oriented anchor box mechanism to improve the accuracy of grasp detection and employed an end-to-end fully convolutional neural network. They utilized only one anchor box with multiple orientations, rather than multiple scales or aspect ratios~\cite{2017AHybridDeep,2018RealWorldMultiObj} for reference grasps, and predicted five regression values and one grasp quality score for each oriented reference box. Depierre et al.~\cite{2020OptimizingCorrelated} further extends Zhou et al.~\cite{2018FullyConvolutionalGrasp} by adding a direct dependency between the regression prediction and the score evaluation. They proposed a novel DNN architecture with a scorer which evaluates the graspability of a given position and introduced a novel loss function which correlates the regression of grasp parameters with the graspability score.

Some other methods are also proposed aiming at cluttered scenes, where a robot need to know if an object is on another object in the piles of objects for a successful grasp. Guo et al.~\cite{2016ObjectDiscovery} presented a shared convolutional neural network to conduct object discovery and grasp detection. Zhang et al.~\cite{2018AMultitaskCNN} proposed a multi-task convolution robotic grasping network to address the problem of combining grasp detection and object detection with relationship reasoning in the piles of objects. The method of Zhang et al.~\cite{2018AMultitaskCNN} consists of several deep neural networks that are responsible for generating local feature maps, grasp detection, object detection and relationship reasoning separately. In comparison, Park et al.~\cite{2019ASingleMultiTask} proposed a single multi-task deep neural networks that yields the information on grasp detection, object detection and relationship reasoning among objects with a simple post-processing.

\begin{table*}[htbp]
\centering
\caption{Summaries of publicly available 2D planar grasp datasets.}
    \begin{tabular}{p{7.0cm}p{2.0cm}<{\centering}p{4.5cm}<{\centering}p{2.5cm}<{\centering}}
        \hline
        Dataset & Objects Num
        &Num of RGB-D images & Num of grasps \\
        \hline
        Stanford Grasping~\cite{2008RoboticGraspingDataset,2008LearningToGrasp} & 10 & 13747 & 13747\\
        Cornell Grasping~\cite{2011EfficientGrasping} & 240 & 885 & 8019\\
        CMU dataset~\cite{2016SupersizingSelfSuper} & over 150 & 50567 & no\\
        Dex-Net 2.0~\cite{2017DexNet2} & over 150 & 6.7 M(Depth only) & 6.7 M\\
        JACQUARD~\cite{2018JacquardDataset} & 11619 & 54485 & 1.1 M\\
        \hline
    \end{tabular}
\label{tab:datasetgrasp}
\end{table*}

\begin{table*}[htbp]
\centering
\caption{Accuracies of grasp prediction on the Cornell Grasp dataset.}
 \begin{tabular}{p{7.0cm}p{3.0cm}<{\centering}p{1.5cm}<{\centering}p{1.5cm}<{\centering}p{2.0cm}<{\centering}}
    \hline
    \multirow{2}{*}{Method} & \multirow{2}{*}{Input Size}  & \multicolumn{2}{c}{Accuracy(\%)} & \multirow{2}{*}{Time} \\
    \cline{3-4}
     & & \multicolumn{1}{c}{Image Split} & \multicolumn{1}{c}{Object Split} & \\
    \hline
    Jiang et al.~\cite{2011EfficientGrasping} & 227 x 227 & 60.50 & 58.30 & 50sec\\
    Lenz et al.~\cite{2015DeepLearnDetect} & 227 x 227 & 73.90 & 75.60 & 13.5sec\\
    Morrison et al.~\cite{2018ClosingTheLoop} & 300 x 300 & 78.56 & - & 7ms\\
    Redmon et al.~\cite{2015RealGrasp} & 224 x 224 & 88.00 & 87.1 & 76ms\\
    Zhang et al.~\cite{2017RobustRobotGrasp} & 224 x 224 & 88.90 & 88.20 & 117ms\\
    Kumra et al.~\cite{2017RoboticGrasp} & 224 x 224 & 89.21 & 88.96 & 103ms\\
    Chun et al.~\cite{2018ClassificationBasedGrasp} & 400 x 400 & 89.60 & - & 23ms\\
    Asif et al.~\cite{2018GraspNet} & 224 x 224 & 90.60 & 90.20 & 24ms\\
    Wang et al.~\cite{2019EfficientFCNN} & 400 x 400 & 94.42 & 91.02 & 8ms\\
    Chu et al.~\cite{2018RealWorldMultiObj} & 227 x 227 & 96.00 & 96.10 & 120ms\\
    Chun et al.~\cite{2018RealTimeHighly} & 360 x 360 & 96.60 & 95.40 & 20ms\\
    Zhou et al.~\cite{2018FullyConvolutionalGrasp} & 320 x 320 & 97.74 & 96.61 & 118ms\\
    Park et al.~\cite{2019ASingleMultiTask} & 360 x 360 & 98.6 & 97.2 & 16ms\\
    \hline
 \end{tabular}
\label{tab:graspcompCornell}
\end{table*}

\subsubsection{Comparisons and Discussions}
\label{4-1-3}

The methods of 2D planar grasp are evaluated in this section, which contain the datasets, evaluation metrics and comparisons of the recent methods.

\paragraph{Datasets and evaluation metrics}

There exist a few datasets for 2D planar grasp, which are presented in Table~\ref{tab:datasetgrasp}. Among them, the Cornell Grasping dataset~\cite{2011EfficientGrasping} is the most widely used dataset. In addition, the dataset has the image-wise splitting and the object-wise splitting. Image-wise splitting splits images randomly and is used to test how well the method can generalize to new positions for objects it has seen previously. Object-wise splitting puts all images of the same object into the same cross-validation split and is used to test how well the method can generalize to novel objects.

Aiming at the point-based grasps and the oriented rectangle-based grasps~\cite{2011EfficientGrasping}, there exist two metrics for evaluating the performance of grasp detection: the point metric and the rectangle metric. The former evaluates the distance between predicted grasp center and the ground truth grasp center w.r.t. a threshold value. It has difficulties in determining the distance threshold and does not consider the grasp angle. The latter metric considers a grasp to be correct if the grasp angle is within $30^\circ $ of the ground truth grasp, and the Jaccard index  $J(A,B) = {{\left| {A \cap B} \right|} \mathord{\left/
 {\vphantom {{\left| {A \cap B} \right|} {\left| {A \cup B} \right|}}} \right.
 \kern-\nulldelimiterspace} {\left| {A \cup B} \right|}}$ of the predicted grasp $A$ and the ground truth $B$ is greater than $25\% $.

\paragraph{Comparisons}

The methods of evaluating oriented rectangles are compared in Table~\ref{tab:graspcompCornell} on the widely used Cornell Grasping dataset~\cite{2011EfficientGrasping}. From the table, we can see that the state-of-the-art methods have achieved very high accuracies on this dataset. Recent works~\cite{2020OptimizingCorrelated} began to conduct experiments on the Jacquard Grasp dataset~\cite{2018JacquardDataset} since it has more images and the grasps are more diverse.

\subsection{6DoF Grasp}
\label{4-2}

Methods of 6DoF grasp can be divided into methods based on the partial point cloud and methods based on the complete shape. These methods are summarized in Table~\ref{tab:6d-grasp}.

\begin{table*}[htbp]
\centering
\caption{Summary of 6DoF grasp estimation methods.}
  \begin{tabular}{m{1.5cm} m{2.0cm} m{4.0cm} m{8.5cm}}
        \toprule
        Methods & Descriptions & Traditional methods & Deep learning-based methods \\

        \midrule
        \multirow{2}{1.5cm}{Methods based on the partial point cloud} & Estimate grasp qualities of candidate grasps & Bohg and Kragic~\cite{2010LearningGraspingPoints}, Pas et al.~\cite{2015UsingGeometryToDetect}, Zapata-Impata et al.~\cite{2019FastGeometry} & GPD~\cite{2017GPD}, PointnetGPD~\cite{2019Pointnetgpd}, 6-DoF GraspNet~\cite{20196DoFGraspNet}, S$^{4}$G~\cite{2019S4g}, REGNet~\cite{2020REGNet} \\

         \cline{2-4}
         & Transfer grasps from existing ones & Andrew et al.~\cite{2003AutomaticGraspPlanning}, Nikandrova and Kyrki~\cite{2015CategorybasedTask}, Vahrenkamp et al.~\cite{2016PartBasedGrasp} & Tian et al.~\cite{2018TransferringGrasp}, Dense Object Nets~\cite{2018DenseObjectNets}, DGCM-Net~\cite{2020DGCMNet}\\

        \hline
        \multirow{2}{1.5cm}{Methods based on the complete shape} & Estimate the 6D object pose & Zeng et al.~\cite{2017MultiViewSelf} & Billings and Roberson~\cite{2018SilhoNet} \\

        \cline{2-4}
        & Conduct shape completion & Miller et al.~\cite{2003AutomaticGraspPlanning} & Varley et al.~\cite{2017ShapeCompletion}, Lundell et al.~\cite{2019RobustGraspPlanning}, Watkins-Valls et al.~\cite{2019Multimodel}, Merwe et al.~\cite{2019LearningContinuous3D}, Wang et al.~\cite{20183DShapePerception}, Yan et al.~\cite{2018Learning6DoF}, Yan et al.~\cite{2019DataEfficientLearning}, Tosun et al.~\cite{2020RoboticGraspingThrough}, kPAM-SC~\cite{2019KpamSC}, ClearGrasp~\cite{2019Cleargrasp} \\

    \bottomrule
    \end{tabular}
\label{tab:6d-grasp}
\end{table*}

\subsubsection{Methods based on the partial point cloud}
\label{4-2-1}

\begin{figure}[htbp]
\centering
\includegraphics[scale=0.35]{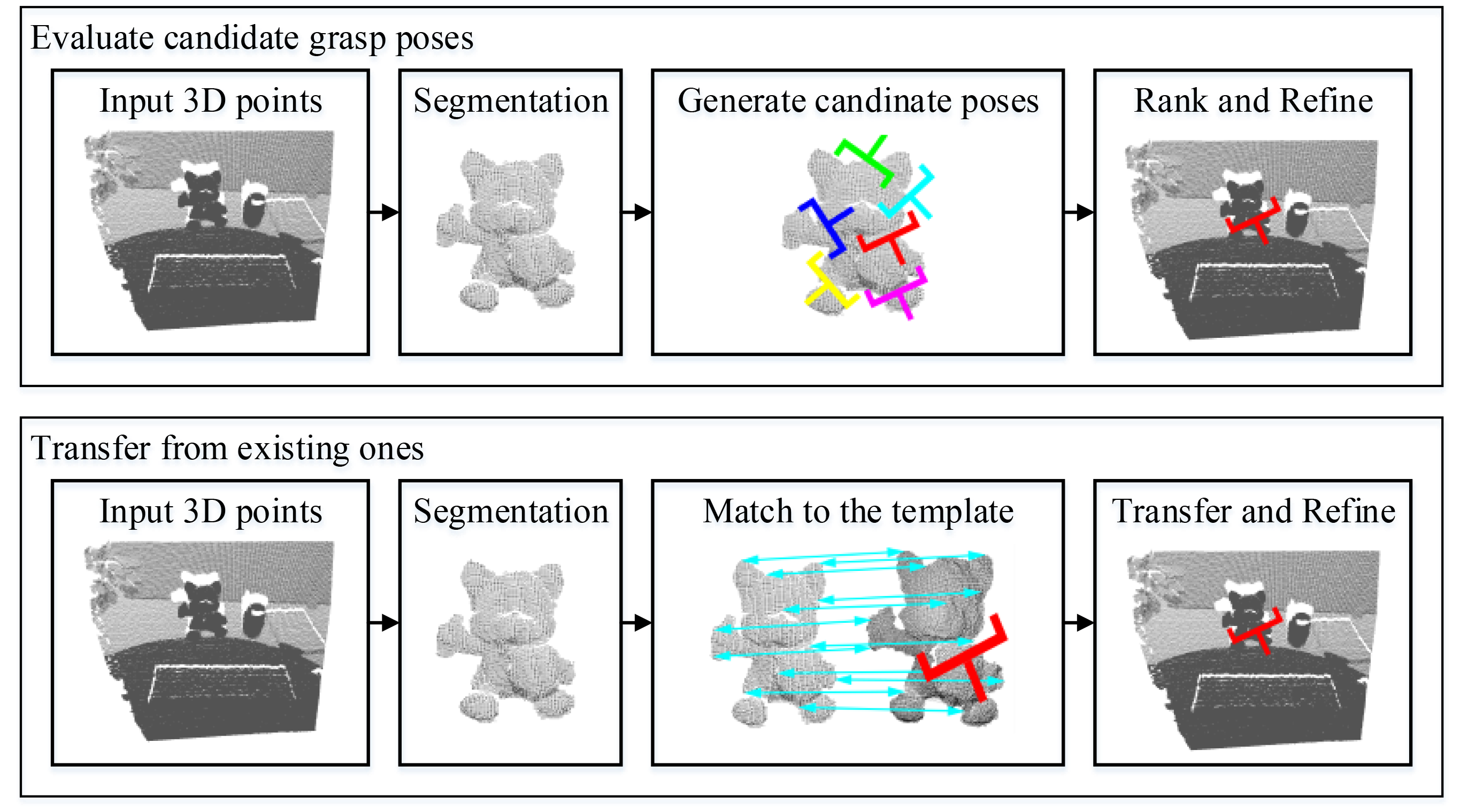}
\caption{Typical functional flow-chart of 6DoF grasp methods based on the partial point cloud.}
\label{fig:6dgrasppartial}
\end{figure}

This kind of methods can be divided into two kinds. The first kind of methods estimate grasp qualities of candidate grasps and the second kind of methods transfer grasps from existing ones. Typical functional flow-chart of methods based on the partial point cloud is illustrated in Fig.~\ref{fig:6dgrasppartial}.

\paragraph{Methods of estimating grasp qualities of candidate grasps}

This kind of methods mean that the 6DoF grasping pose is estimated through analyzing the input partial point cloud merely. Most of this kind of methods~\cite{2010LearningGraspingPoints,2015UsingGeometryToDetect,2019FastGeometry,2017GPD,2019Pointnetgpd} sample large number of candidate grasps first, and then utilize various methods to evaluate grasp qualities, which is a classification-based manner. While some novel methods~\cite{2019S4g,2020REGNet,2020PointnetGrasping,20196DoFGraspNet} estimate the grasp qualities implicitly and directly predict the 6DoF grasp pose in a single-shot way, which is a regression-based manner.

Bohg and Kragic~\cite{2010LearningGraspingPoints} applied the concept of shape context~\cite{2002ShapeContext} to improve the performance of grasping point classification. They used a supervised learning approach and the classifier is trained with labelled synthetic images. Pas et al.~\cite{2015UsingGeometryToDetect} first used a geometrically necessary condition to sample a large set of high quality grasp hypotheses, which will be classified using the notion of an antipodal grasp. Zapata-Impata et al.~\cite{2019FastGeometry} proposed a method to find the best pair of grasping points given a partial single-view point cloud of an unknown object. They defined an improved version of the ranking metric~\cite{2017UsingGeometryTo} for evaluating a pair of contact points, which is parameterized by the morphology of the robotic hand in use.

3D data has different representations such as multi-view images, voxel grids or point cloud, and each representation can be processed with corresponding deep neural networks. These different kinds of neural networks have already been applied into robotic grasping tasks. GPD~\cite{2017GPD} generates candidate grasps on the a region of interest (ROI) first. These candidate grasps are then encoded into a stacked multi-channel image. Each candidate is evaluated to obtain a score using a four-layer convolutional neural network finally. Lou et al.~\cite{2019LearningtoGenerate} proposed an algorithm that uniformly samples over the entire 3D space first to generate candidate grasps, predicts the grasp stability using 3D CNN together with a grasping reachability using the candidate grasp pose, and obtains the final grasping success probability. PointnetGPD~\cite{2019Pointnetgpd} randomly samples candidate grasps, and evaluates the grasp quality by direct point cloud analysis with the 3D deep neural network PointNet~\cite{2017Pointnet}. During the generation of training datasets, the grasp quality is evaluated through combining the force-closure metric and the Grasp Wrench Space (GWS) analysis~\cite{1992Quantitative}. Mousavian et al.~\cite{20196DoFGraspNet} proposed an algorithm called 6-DoF GraspNet, which samples grasp proposals using a variational auto-encoder and refines the sampled grasps using a grasp evaluator model. Pointnet++~\cite{2017Pointnet++} is used to generate and evaluate grasps. Murali et al.~\cite{20196DoFGrasping} further improved 6-DoF GraspNet by introducing a learned collision checker conditioned on the gripper information and on the raw point cloud of the scene, which affords a higher success rate in cluttered scenes.

Qin et al.~\cite{2019S4g} presented an algorithm called S$^{4}$G, which utilizes a single-shot grasp proposal network trained with synthetic data using Pointnet++~\cite{2017Pointnet++} and predicts amodal grasp proposals efficiently and effectively. Each grasp proposal is further evaluated with a robustness score. The core novel insight of S$^{4}$G is that they learn to propose possible grasps by regression, rather than using a sliding windows-like style. S$^{4}$G generates grasp proposals directly, while 6-DoF GraspNet uses an encode and decode way. Ni et al.~\cite{2020PointnetGrasping} proposed Pointnet++Grasping, which is also an end-to-end approach to directly predict the poses, categories and scores of all the grasps. Further, Zhao et al.~\cite{2020REGNet} proposed an end-to-end single-shot grasp detection network called REGNet, which takes one single-view point cloud as input for parallel grippers. There network contains three stages, which are the Score Network (SN) to select positive points with high grasp confidence, the Grasp Region Network (GRN) to generate a set of grasp proposals on selected positive points, and the Refine Network (RN) to refine the detected grasps based on local grasp features. REGNet is the state-of-the-art method for grasp detection in 3D space and outperforms several methods including GPD~\cite{2017GPD}, PointnetGPD~\cite{2019Pointnetgpd} and S$^{4}$G~\cite{2019S4g}. Fang et al.~\cite{2020GraspNet1Billion} proposed a large-scale grasp pose detection dataset called GraspNet-1Billion, which contains 97,280 RGB-D image with over one billion grasp poses. They also proposed an end-to-end grasp pose prediction network that learns approaching direction and operation parameters in a decoupled manner.

\paragraph{Methods of transferring grasps from existing ones}

This kind of methods transfer grasps from existing ones, which means finding correspondences from the observed single-view point cloud to the existing complete one if we know that they come from one category. In most cases, target objects are not totally the same with the objects in the existing database. If an object comes from a class that is involved in the database, it is regarded as a similar object. After the localization of the target object, correspondence-based methods can be utilized to transfer the grasp points from the similar and complete 3D object to the current partial-view object. These methods learn grasps by observing the object without estimating its 6D pose, since the current target object is not totally the same with the objects in the database.

Different kinds of methods are utilized to find the correspondences based on taxonomy, segmentation, and so on. Andrew et al.~\cite{2003AutomaticGraspPlanning} proposed a taxonomy-based approach, which classified objects into categories that should be grasped by each canonical grasp. Nikandrova and Kyrki~\cite{2015CategorybasedTask} presented a probabilistic approach for task-specific stable grasping of objects with shape variations inside the category. An optimal grasp is found as a grasp that is maximally likely to be task compatible and stable taking into account shape uncertainty in a probabilistic context. Their method requires partial models of new objects, and few models and example grasps are used during the training. Vahrenkamp et al.~\cite{2016PartBasedGrasp} presented a part-based grasp planning approach to generate grasps that are applicable to multiple familiar objects. The object models are segmented according to their shape and volumetric information, and the objet parts are labeled with semantic and grasping information. A grasp transferability measure is proposed to evaluate how successful planned grasps are applied to novel object instances of the same object category. Tian et al.~\cite{2018TransferringGrasp} proposed a method to transfer grasp configurations from prior example objects to novel objets, which assumes that the novel and example objects have the same topology and similar shapes. They perform 3D segmentation on the objects considering geometric and semantic shape characteristics, compute a grasp space for each part of the example object using active learning, and build bijective contact mappings between the model parts and the corresponding grasps for novel objects. Florence et al.~\cite{2018DenseObjectNets} proposed Dense Object Nets, which is built on self-supervised dense descriptor learning and takes dense descriptors as a representation for robotic manipulation. They could grasp specific points on objects across potentially deformed configurations, grasp objects with instance-specificity in clutter, or transfer specific grasps across objects in class. Patten et al.~\cite{2020DGCMNet} presented DGCM-Net, a dense geometrical correspondence matching network for incremental experience-based robotic grasping. They apply metric learning to encode objects with similar geometry nearby in feature space, and retrieve relevant experience for an unseen object through a nearest neighbour search. DGCM-Net also reconstructs 3D-3D correspondences using the view-dependent normalized object coordinate space to transform grasp configurations from retrieved samples to unseen objects. Their method could be extended for semantic grasping by guiding grasp selection to the parts of objects that are relevant to the object's functional use.

\paragraph{Comparisons and discussions}

Methods of estimating grasp qualities of candidate grasps gain much attentions since this is the direct manner to obtain the 6D grasp pose. Aiming at 6DoF grasp, the evaluation metrics for 2D planar grasp are not suitable. The commonly used metric is the Valid Grasp Ratio (VGR) proposed by REGNet~\cite{2020REGNet}. VGR is defined as the quotient of antipodal and collision-free grasps and all grasps. The usually used grasp dataset for evaluation is the YCB-Video~\cite{2018Posecnn} dataset. Comparisons with recent methods are shown in Table~\ref{tab:graspcompYCB}.

\begin{table}[htbp]
\centering
\caption{Accuracies of grasp prediction on the Cornell Grasp dataset.}
 \begin{tabular}{m{4.0cm}m{1.5cm}<{\centering}m{1.5cm}<{\centering}}
    \hline
     Method & VGR(\%) & Time(ms) \\
    \hline
    GPD~\cite{2017GPD} (3 channels) & 79.34 & 2077.12\\
    GPD~\cite{2017GPD} (12 channels) & 80.22 & 2702.38\\
    PointNetGPD~\cite{2019Pointnetgpd} & 81.48 & 1965.60\\
    S$^{4}$G~\cite{2019S4g} & 77.63 & 679.04\\
    REGNet~\cite{2020REGNet} & 92.47 & 686.31\\
    \hline
 \end{tabular}
\label{tab:graspcompYCB}
\end{table}

Methods of transferring grasps from existing ones have potential usages in high-level robotic manipulation tasks. Not only the grasps could be transferred, the manipulation skills could also be transferred. Lots of methods~\cite{2019RobotLearningOfShifting,2019LearningActionsFromHD} that learn grasps from demonstration usually utilize this kind of methods.

\subsubsection{Methods based on the complete shape}
\label{4-2-2}

\begin{figure}[htbp]
\centering
\includegraphics[scale=0.35]{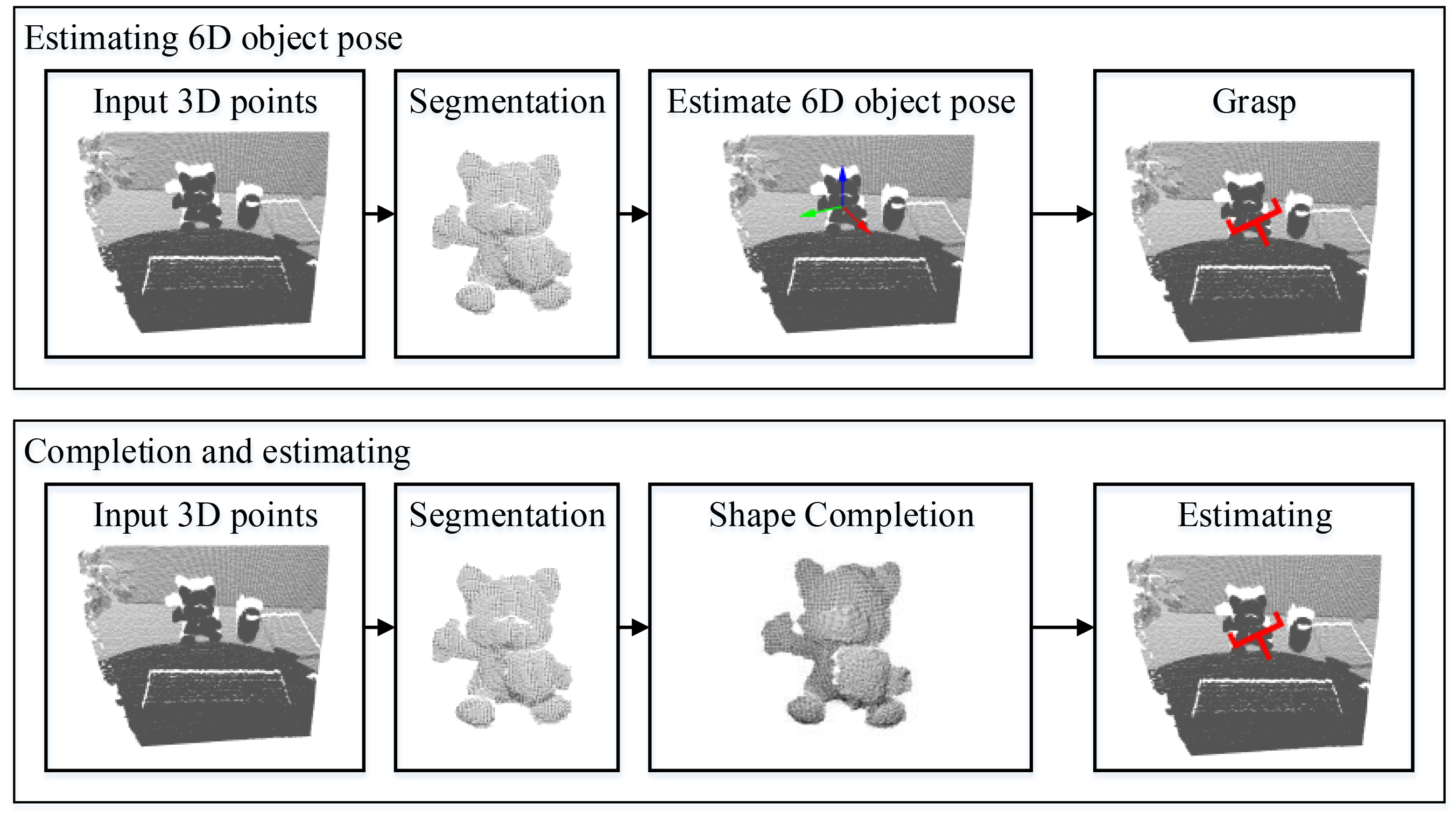}
\caption{Typical functional flow-chart of 6DoF grasp methods based on the complete shape.}
\label{fig:6dgraspcomplete}
\end{figure}

Methods based on the partial point cloud are suitable for unknown objects, since these methods have no identical 3D models to use. Aiming at known objects, their 6D poses can be estimated and the 6DoF grasp poses estimated on the complete 3D shape could be transformed from the object coordinate to the camera coordinate. In another perspective, the 3D complete object shape under the camera coordinate could also be completed from the observed single-view point cloud. And the 6DoF grasp poses could be estimated based on the completed 3D object shape in the camera coordinate. We consider these two kinds of methods as complete shape-based methods since 6DoF grasp poses are estimated based on complete object shapes. Typical functional flow-chart of 6DoF grasp methods based on the complete shape is illustrated in Fig.~\ref{fig:6dgraspcomplete}.

\paragraph{Methods of estimating the 6D object pose}

The 6D object pose could be accurately estimated from the RGB-D data if the target object in known as mentioned in Section~\ref{sec:3}, and 6DoF grasp poses can be obtained via offline pre-computation or online generation. This is the most popular method used for the grasping systems. If the 6DoF grasp poses exist in the database, the current 6DoF grasp pose could be retrieved from the knowledge base, or obtained by sampling and ranking them through comparisons with existing grasps. If the 6DoF grasp poses do not exist in the database, analytical methods are utilized to compute the grasp poses. Analytical methods consider kinematics and dynamics formulation in determining grasps~\cite{2012GraspSurvey}. Force-closure is one of the main conditions in completing the grasping tasks and there exist many force-closure grasp synthesis methods for 3D objects. Among them, the polyhedral objects are first dealt with, as they are composed of a finite number of flat faces. The force-closure condition is reduced into the test of the angles between the faces normals~\cite{1987Constructing} or using the linear model to derive analytical formulation for grasp characterization~\cite{1993OnCharacterizing}. To handle the commonly used objects which usually have more complicated shapes, methods of observing different contact points are proposed~\cite{2001OnComputingImm}. These methods try to find contact points on a 3D object surface to ensure force-closure and compute the optimal grasp by minimizing an objective energy function according to a predefined grasp quality criterion~\cite{1994EasilyComputable}. However, searching the grasp solution space is a complex problem which is quite time-consuming. Some heuristical techniques were then proposed to reduce the search space by generating a set of grasp candidates according to a predefined procedure~\cite{2003GraspingTheDice}, or by defining a set of rules to generate the starting positions~\cite{2003AutomaticGraspPlanning}. A few robotic grasping simulators, such as GraspIt!~\cite{2004GraspIt}, assist the generation of the best gripper pose to conduct a successful grasp. Andrew and Peter~\cite{2004GraspIt} proposed GraspIt!, which is a versatile simulator for robotic grasping. GraspIt! supports the loading of objects and obstacles of arbitrary geometry to populate a complete simulation world. It allows a user to interactively manipulate a robot or an object and create contacts between them. Xue et al.~\cite{2009AutomaticGraspPlan} implemented a grasping planning system based on GraspIt! to plan high-quality grasps. Le{\'o}n et al.~\cite{2010OpenGRASP} presented OpenGRASP, a toolkit for simulating grasping and dexterous manipulation. It provides a holistic environment that can deal with a variety of factors associated with robotic grasping. These methods produce successful grasps and detailed reviews could be found in the survey~\cite{2012GraspSurvey}.

Both traditional and deep learning-based 6D object pose estimation algorithms are utilized to assist the robotic grasping tasks. Most of the methods~\cite{2017MultiViewSelf} presented in the Amazon picking challenge utilize the 6D poses estimated through partial registration first. Zeng et al.~\cite{2017MultiViewSelf} proposed an approach which segments and labels multiple views of a scene with a fully convolutional neural network, and then fits pre-scanned 3D object models to the segmentation results to obtain the 6D object poses. Besides, Billings and Johnson-Roberson~\cite{2018SilhoNet} proposed a method which jointly accomplish object pose estimation and grasp point selection using a Convolutional Neural Network (CNN) pipeline. Wong et al.~\cite{2017Segicp} proposed a method which integrated RGB-based object segmentation and depth image-based partial registration to obtain the pose of the target object. They presented a novel metric for scoring model registration quality, and conducted multi-hypothesis registration, which achieved accurate pose estimation with $1 cm$ position error and $< {5^\circ }$ angle error. Using this accurate 6D object pose, grasps are conducted with a high success rate. A few deep learning-based 6D object pose estimation approaches such as DenseFusion~\cite{2019DenseFusion} also illustrate high successful rates in conducting practical robotic grasping tasks.

\paragraph{Methods of conducting shape completion}

There also exist one kind of methods, which conduct 3D shape completion for the partial point cloud, and then estimate grasps. 3D shape completion provides the complete geometry of objects from partial observations, and estimating 6DoF grasp poses on the completed shape is more precise. Most of this kind of methods estimate the object geometry from partial point cloud~\cite{2017ShapeCompletion,2019RobustGraspPlanning,2019LearningContinuous3D,2019Multimodel,2020RoboticGraspingThrough}, and some other methods~\cite{20183DShapePerception,2018Learning6DoF,2019DataEfficientLearning,2019KpamSC,2019Cleargrasp} utilize the RGB-D images. Many of them~\cite{20183DShapePerception,2019Multimodel} also combine tactile information for better prediction.

Varley et al.~\cite{2017ShapeCompletion} proposed an architecture to enable robotic grasp planning via shape completion. They utilized a 3D convolutional neural network (CNN) to complete the shape, and created a fast mesh for objects not to be grasped, a detailed mesh for objects to be grasped. The grasps are finally estimated on the reconstructed mesh in GraspIt!~\cite{2004GraspIt} and the grasp with the highest quality is executed. Lundell et al.~\cite{2019RobustGraspPlanning} proposed a shape completion DNN architecture to capture shape uncertainties, and a probabilistic grasp planning method which utilizes the shape uncertainty to propose robust grasps. Merwe et al.~\cite{2019LearningContinuous3D} proposed PointSDF to learn a signed distance function implicit surface for a partially viewed object, and proposed a grasp success prediction learning architecture which implicitly learns geometrically aware point cloud encodings. Watkins-Valls et al.~\cite{2019Multimodel} also incorporated depth and tactile information to create rich and accurate 3D models useful for robotic manipulation tasks. They utilized both the depth and tactile as input and fed them directly into the model rather than using the tactile information to refine the results. Tosun et al.~\cite{2020RoboticGraspingThrough} utilized a grasp proposal network and a learned 3D shape reconstruction network, where candidate grasps generated from the first network are refined using the 3D reconstruction result of the second network. These above methods mainly utilize depth data or point cloud as inputs.

Wang et al.~\cite{20183DShapePerception} perceived accurate 3D object shape by incorporating visual and tactile observations, as well as prior knowledge of common object shapes learned from large-scale shape repositories. They first applied neural networks with learned shape priors to predict an object's 3D shape from a single-view color image and the tactile sensing was used to refine the shape. Yan et al.~\cite{2018Learning6DoF} proposed a deep geometry-aware grasping network (DGGN), which first learn a 6DoF grasp from RGB-D input. DGGN has a shape generation network and an outcome prediction network. Yan et al.~\cite{2019DataEfficientLearning} further presented a self-supervised shape prediction framework that reconstructs full 3D point clouds as representation for robotic applications. They first used an object detection network to obtain object-centric color, depth and mask images, which will be used to generate a 3D point cloud of the detected object. A grasping critic network is then used to predict a grasp. Gao and Tedrake~\cite{2019KpamSC} proposed a new hybrid object representation consisting of semantic keypoints and dense geometry (a point cloud or mesh) as the interface between the perception module and motion planner. Leveraging advances in learning-based keypoint detection and shape completion, both dense geometry and keypoints can be perceived from raw sensor input. Sajjan et al.~\cite{2019Cleargrasp} presented ClearGrasp, a deep learning approach for estimating accurate 3D geometry of transparent objects from a single RGB-D image for robotic manipulation. ClearGrasp uses deep convolutional networks to infer surface normals, masks of transparent surfaces, and occlusion boundaries, which will refine the initial depth estimates for all transparent surfaces in the scene.

\paragraph{Comparisons and Discussions}
\label{sec:4-2-3}

When accurate 3D models are available, the 6D object pose could be achieved, which affords the generation of grasps for the target object. However, when existing 3D models are different from the target one, the 6D poses will have a large deviation, and this will lead to the failure of the grasp. In this case, we can complete the partial-view point cloud and triangulate it to obtain the complete shape. The grasps could be generated on the reconstructed and complete 3D shape. Various grasp simulation toolkits are developed to facilitate the grasps generation.

Aiming at methods of estimating the 6D object pose, there exist some challenges. Firstly, this kind of methods highly rely on the accuracy of object segmentation. However, training a network which supports a wide range of objects is not easy. Meanwhile, these methods require the 3D object to grasp be similar enough to those of the annotated models such that correspondences can be found. It is also challenging to compute grasp points with high qualities for objects in cluttered environments where occlusion usually occurs. Aiming at methods of conducting shape completion, there also exist some challenges. The lack of information, especially the geometry on the opposite direction from the camera, extremely affect the completion accuracy. However, using multi-source data would be a future direction.

\section{Challenges and Future Directions}
\label{sec:5}

In this survey, we review related works on vision-based robotic grasping from three key aspects: object localization, object pose estimation and grasp estimation. The purpose of this survey is to allow readers to get a comprehensive map about how to detect a successful grasp given the initial raw data. Various subdivided methods are introduced in each section, as well as the related datasets and comparisons. Comparing with existing literatures, we present an end-to-end review about how to conduct a vision-based robotic grasp detection system.

Although so many intelligent algorithms are proposed to assist the robotic grasping tasks, challenges still exist in practical applications, such as the insufficient information in data acquisition, the insufficient amounts of training data, the generalities in grasping novel objects and the difficulties in grasping transparent objects.

The first challenge is the insufficient information in data acquisition. Currently, the mostly used input to decide a grasp is one RGB-D image from one fixed position, which lacks the information backwards. It's really hard to decide the grasp when we do not have the full object geometry. Aiming at this challenge, some strategies could be adopted. The first strategy is to utilize multi-view data. A more widely perspective data is much better since the partial views are not enough to get a comprehensive knowledge of the target object. Methods based on poses of the robotic arms~\cite{2017MultiViewSelf,2020FoFetch} or the slam methods~\cite{2017BundleFusion} can be adopted to merge the multi-view data. Instead of fusing multi-view data, the best grasping view could also be chosen explicitly~\cite{2019MultiViewPicking}. The second one is to involve multi-sensor data such as the haptic information. There exist some works~\cite{2019MakingSenseVisionTouch,2019TransferLearingHaptic,2020TactileDexterity} which already involve the tactile data to assist the robotic grasping tasks.

The second challenge is the insufficient amounts of training data. The requirements for the training data is extremely large if we want to build an intelligent enough grasp detection system. The amount of open grasp datasets is really small and the involved objects are mostly instance-level, which is too small compared with the objects in our daily life. Aiming at this challenges, some strategies could be adopted. The first strategy is to utilize simulated environments to generate virtual data~\cite{2018DeepObjectPoseEst}. Once the virtual grasp environments are built, large amounts of virtual data could be generated by simulating the sensors from various angles. Since there exists gaps from the simulation data to the practical one, many domain adaptation methods~\cite{2018UsingSimulationandDA,2018MultiTaskDA,2020MultiSourceDA} have been proposed. The second strategy is to utilize the semi-supervised learning approaches~\cite{2020SemiSupervisedGrasp,2020MultiLearningFewShot} to learn to grasp with incorporate unlabeled data. The third strategy is to utilize self-supervised learning methods to generate the labeled data for 6D object pose estimation~\cite{2020SelfSupervised6D} or grasp detection~\cite{2020OnlineSelfSupervised}.

The third challenge is the generalities in grasping novel objects. The mentioned grasp estimation methods, except for methods of evaluating the 6D object pose, all have certain generalities in dealing with novel objects. But these methods mostly work well on trained dataset and show reduced performance for novel objects. Other than improving the performance of the mentioned algorithms, some strategies could be adopted. The first strategy is to utilize the category-level 6D object pose estimation. Lots of works~\cite{2019NOCS,2020LatentFusion,20196-PACK,2020LearningCASS} start to deal with the 6D object pose estimation of category-level objects, since high performance have been achieved on instance-level objects. The second strategy is to involve more semantic information in the grasp detection system. With the help of various shape segmentation methods~\cite{2019PartNetShapeSeg,2020LearningToGroup}, parts of the object instead of the complete shape can be used to decrease the range of candidate grasping points. The surface material and the weight information could also be estimated to obtain more precise grasping detection results.

The fourth challenge lies in grasping transparent objects. Transparent objects are prevalent in our daily life, but capturing their 3D information is rather difficult for nowadays depth sensors. There exist some pioneering works that tackle this problem in different ways. GlassLoc~\cite{2019GlassLoc} was proposed for grasp pose detection of transparent objects in transparent clutter using plenoptic sensing. KeyPose~\cite{2020KeyPose} conducted multi-view 3D labeling and keypoint estimation for transparent objects in order to estimate their 6D poses. ClearGrasp~\cite{2019Cleargrasp} estimates accurate 3D geometry of transparent objects from a single RGB-D image for robotic manipulation. This area will be further researched in order to make grasps much accurate and robust in daily life.

%

\bibliographystyle{named}
\bibliography{grasp}

\end{document}